\pdfoutput=1
\documentclass[11pt]{article}

\usepackage[]{acl}

\usepackage{times}
\usepackage{latexsym}
\usepackage[utf8]{inputenc}
\usepackage{textgreek}
\usepackage[T1]{fontenc}
\usepackage{CJKutf8}
\usepackage[english]{babel}
\usepackage{makecell}
\usepackage{graphicx}
\usepackage{hyperref}
\usepackage{multirow} % Table 기능
\usepackage{float} % H
\usepackage{amsmath} % cases / align
\usepackage{adjustbox} % \resizebox{\textwidth}{!}
\usepackage{verbatim} % comment 기능
\usepackage{hwemoji}
\usepackage{pifont} % emoji 
\usepackage[normalem]{ulem}
\useunder{\uline}{\ul}{}
\usepackage{array}
\usepackage[T1]{fontenc}
\usepackage[utf8]{inputenc}

\usepackage{microtype}

\newcommand\blfootnote[1]{%
  \begingroup
  \renewcommand\thefootnote{}\footnote{#1}%
  \addtocounter{footnote}{-1}%
  \endgroup
}

\title{Plug-in and Fine-tuning: Bridging the Gap between Small Language Models and Large Language Models}

\author{Kyeonghyun Kim\textsuperscript{1*}, Jinhee Jang\textsuperscript{1*}, Juhwan Choi\textsuperscript{2*†}, \\ 
\textbf{Yoonji Lee\textsuperscript{1}, Kyohoon Jin\textsuperscript{3†}, YoungBin Kim\textsuperscript{1}}  \\
  \textsuperscript{1}Chung-Ang University \quad
  \textsuperscript{2}AITRICS \quad
  \textsuperscript{3}DATUMO \quad \\
  \textsuperscript{1}\texttt{\{khyun8072, jinheejang, pioneer0305, ybkim85\}@cau.ac.kr}  \\
  \textsuperscript{2}\texttt{jhchoi@aitrics.com} \quad \textsuperscript{3}\texttt{kyohoon.jin@selectstar.ai}
}

\begin{document}
\maketitle

\blfootnote{\textsuperscript{*}Equal contribution.}
\blfootnote{\textsuperscript{†}Work was done at Chung-Ang University.}

\begin{abstract}
Large language models (LLMs) are renowned for their extensive linguistic knowledge and strong generalization capabilities, but their high computational demands make them unsuitable for resource-constrained environments. In contrast, small language models (SLMs) are computationally efficient but often lack the broad generalization capacity of LLMs. To bridge this gap, we propose PiFi, a novel framework that combines the strengths of both LLMs and SLMs to achieve high performance while maintaining efficiency. PiFi integrates a single frozen layer from an LLM into a SLM and fine-tunes the combined model for specific tasks, boosting performance without a significant increase in computational cost. We show that PiFi delivers consistent performance improvements across a range of natural language processing tasks, including both natural language understanding and generation. Moreover, our findings demonstrate PiFi’s ability to effectively leverage LLM knowledge, enhancing generalization to unseen domains and facilitating the transfer of linguistic abilities.\end{abstract}

\section{Introduction}
\label{sec:intro}
Language models (LMs) based on transformer architecture have demonstrated impressive performance across a wide range of natural language processing (NLP) tasks, primarily due to the linguistic knowledge they acquire from training on large-scale datasets \cite{zhao2023survey}. In particular, large language models (LLMs), such as GPT-3 \cite{brown2020language}, GPT-4 \cite{achiam2023gpt}, and Llama-3 \cite{dubey2024llama}, stand out for their significantly larger number of parameters compared to small language models (SLMs), which typically contain between 100M and 5B parameters \cite{lu2024small}. Notable examples of SLMs include BERT \cite{devlin2019bert}, RoBERTa \cite{liu2019roberta}, and BART \cite{lewis2020bart}. The extensive parameterization of LLMs allows them to capture a broader range of knowledge, leading to enhanced generalizability across novel tasks, whereas SLMs often require fine-tuning for specific tasks or domains \cite{brown2020language, wei2022emergent, ye2024cross}.

Despite the notable strengths and exceptional performance of LLMs, they come with inherent limitations. A major constraint is the trade-off between performance and inference cost, primarily due to the computational resources required to deploy these large models \cite{shashidhar2023democratizing}. As the number of parameters increases, so does the demand for computational power and memory, making LLMs less suitable for environments with limited resources, such as mobile devices \cite{nityasya2020costs,lin2023pushing}. Similarly, fine-tuning LLMs for domain-specific applications involves a substantial computational overhead \cite{dettmers2023qlora,zhang2024llamaadapter}.

In such resource-constrained scenarios, SLMs often provide a more viable solution. Studies have shown that fine-tuned SLMs can outperform LLMs in specific tasks such as sentiment analysis, semantic textual similarity evaluation, and named entity recognition \cite{yu2023open, lepagnol2024small}. Their lightweight architecture makes them an appealing choice for scenarios where computational efficiency and reduced memory usage are critical \cite{gao2023small, lepagnol2024small}.

To address these challenges, we propose the plug-in and fine-tuning (PiFi) framework, which aims to harness the extensive knowledge and strengths of LLMs, such as linguistic ability and domain generalizability, while retaining the computational efficiency of SLMs. PiFi achieves this by extracting a single layer from a designated LLM and integrating it into a SLM, followed by fine-tuning the combined model on the target task. By incorporating a single LLM layer rather than the full model, PiFi enhances the SLM without compromising its lightweight structure. Furthermore, PiFi optimizes fine-tuning by freezing the extracted LLM layer, thereby minimizing the number of additional parameters to be trained.

To validate the effectiveness of the PiFi framework, we conducted extensive experiments across various datasets, involving diverse natural language understanding (NLU) and natural language generation (NLG) tasks. Additionally, our evaluations in settings such as domain adaptation and multilingual classification demonstrate that PiFi can impart LLMs’ benefits to SLMs, such as improved domain generalizability. In particular, our multilingual classification experiments show that PiFi can significantly enhance SLMs performance by leveraging a layer from an LLM pre-trained in the desired language, illustrating that even a straightforward integration of a single LLM layer can substantially assist SLM training through the knowledge transfer from the LLM. We also conducted a comprehensive comparison of different LLMs, evaluated the impact of varying model sizes, and assessed the effect of integrating instruction-tuned LLMs, highlighting the versatility and robustness of the PiFi framework.

\begin{figure*}[t]
    \centering
    \includegraphics[width=0.8\textwidth]{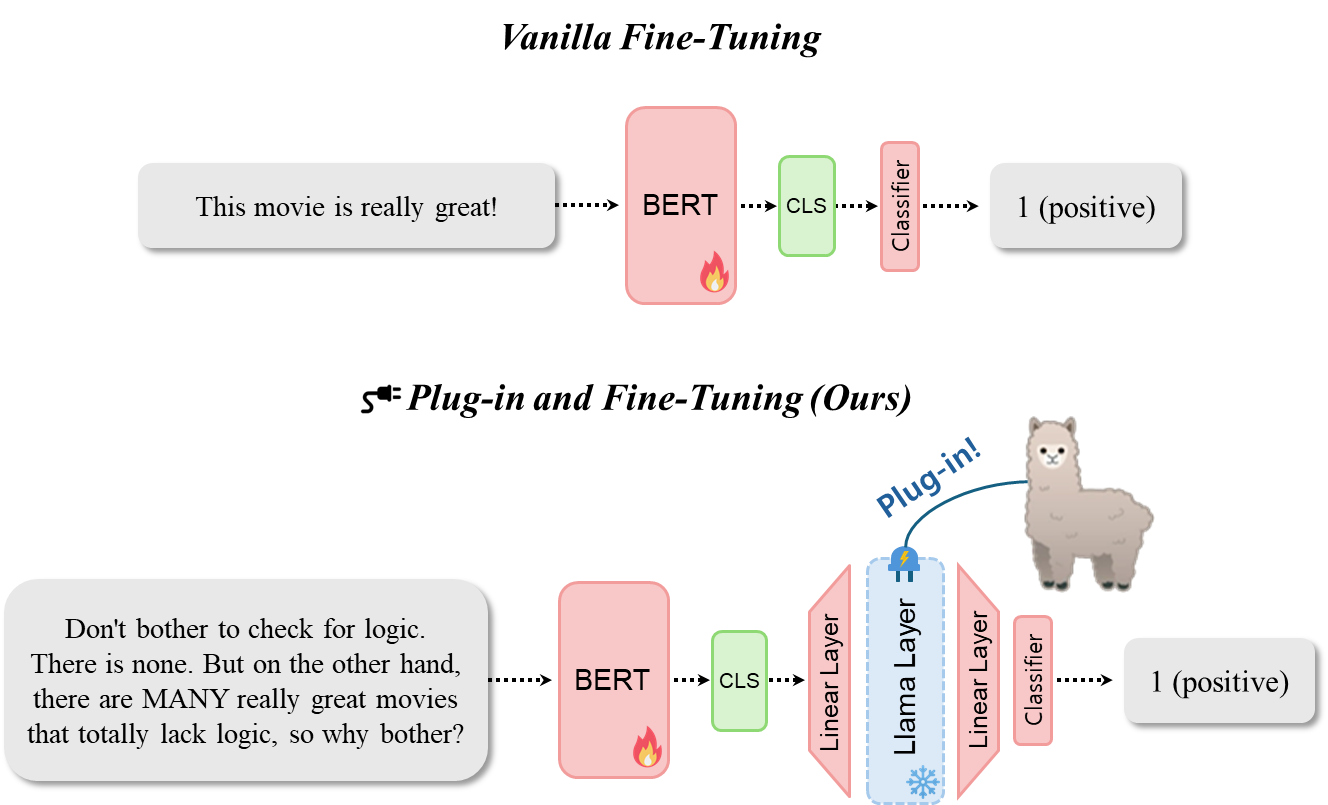}
    \caption{The comparison between vanilla fine-tuning of SLMs and our proposed PiFi architecture.}
\label{fig:framework}
\end{figure*}

\section{Related Work}
\label{sec:related_work}
\subsection{Small Language Models}

Language models (LMs) have progressed significantly from early models like BERT \cite{devlin2019bert} and GPT-1 \cite{Radford2018ImprovingLU}, both of which are based on the transformer architecture \cite{vaswani2017attention}, to more advanced models such as RoBERTa \cite{liu2019roberta}, DeBERTa DeBERTa \cite{he2021deberta}, and BART \cite{lewis2020bart}. These models utilize larger datasets and advanced strategies, achieving greater performance. With each iteration, these models have enhanced their performance and applicability across diverse tasks and languages.

Currently, open-sourced LMs are available in a range of parameter sizes, from extremely small models with 14.5 million parameters \cite{jiao2020tinybert} to large models with billions of parameters \cite{jiang2023mistral, almazrouei2023falcon, dubey2024llama, yang2024qwen2, team2024gemma}. In this study, we focus on specifically on SLMs with millions of parameters. Recent research has introduced several strategies to boost the performance and expand the utility of SLMs while maintaining computational efficiency \cite{gururangan2020dont, gao2023small}. Building upon these works, we propose the PiFi framework, which leverages the strengths of LLMs to supplement SLMs, effectively bridging the gap between SLMs and their larger counterparts.

\subsection{Employment of Large Language Models}

LLMs have achieved state-of-the-art performance across a broad range of NLP tasks, owing to their vast number of parameters. Models like GPT-4 \cite{achiam2023gpt}, Llama-3 \cite{dubey2024llama}, and Mistral \cite{jiang2023mistral} excel in generative tasks, particularly under zero-shot and few-shot settings. Additionally, LLMs demonstrate strong domain generalizability, making them applicable to a variety of domains without additional fine-tuning \cite{minaee2024large}.

These strengths have inspired researchers to explore novel strategies for leveraging LLMs beyond simple downstream task applications. One prominent approach is knowledge distillation, where the knowledge of LLMs is transferred to smaller models. There are two main categories of knowledge distillation methods: parametric and non-parametric. Parametric methods involve using white-box LLMs as teacher models and training student models using the output distribution or intermediate features of the teacher model \cite{zhong2024seeking, timiryasov-tastet-2023-baby, gu2024minillm}. Non-parametric methods, on the other hand, generate synthetic training data using LLMs, which is then used to train smaller student models, thereby achieving knowledge distillation from a data-centric perspective \cite{ye2022zerogen, west2022symbolic, gao2023self, choi2024unigen}.

In recent years, LLMs have also been integrated into multi-modal tasks, extending their utility beyond traditional NLP applications. For instance, some studies have used LLMs’ linguistic capabilities to enhance the pre-training of vision-language models such as CLIP \cite{radford2021learning}, by rewriting the textual descriptions in image-text pairs to improve the model’s understanding of the relationships between images and text \cite{fan2023improving}. Other research has explored using LLMs in computer vision, demonstrating that a combination of a vision transformer (ViT) model \cite{dosovitskiy2021an} with an LLM layer can enhance performance on image classification and other vision tasks \cite{pang2024frozen}. The study showed that LLM layers can serve as visual encoders, helping to identify patterns among image tokens and highlight salient features.

Inspired by these findings, we extend the use of LLMs to various NLP tasks. Specifically, we investigate whether key strengths of LLMs—such as generalizability to unseen domains and their rich linguistic knowledge—can be effectively distilled into SLMs, enabling them to benefit from these capabilities. Our work thus aims to bridge the performance gap between small and large models by transferring these strengths through a novel integration and fine-tuning framework.

\section{Method}
\label{sec:method}
This section presents the PiFi framework, which integrates a single layer from an LLM into SLMs to leverage the extensive knowledge of LLMs while maintaining the efficiency of smaller models. We first describe the methodology for incorporating an LLM layer into encoder-based LMs, such as BERT, and encoder-decoder LMs like T5 \cite{raffel2020exploring}. Subsequently, we detail the fine-tuning process for SLMs with the integrated LLM layer. The overall procedure of the PiFi framework is illustrated in Figure~\ref{fig:framework}.
  
\subsection{Plug-in of LLM Layer to Encoder-based LMs}

An encoder-based model, denoted as $\textit{Enc}$, generates a hidden representation $h_{\textit{enc}}$ for an input sequence $x$, which is then passed to a classification head $\textit{Head}$ to produce the final prediction $\hat{y}$. This process is formulated as:

\begin{gather*}
h_{\textit{enc}} = \textit{Enc}(x), \hat{y} = \textit{Head}(h_{\textit{enc}})
\end{gather*}

To extend this process, PiFi introduces a single LLM layer, denoted as $L_{\textit{LLM}}$, into the pipeline between $\textit{Enc}$ and $\textit{Head}$. Since the hidden representation size of SLMs (e.g., 768 for BERT) may differ from that of LLMs (e.g., 4096 for Llama-3), we employ two additional transformation layers: $L_{\textit{in}}$ and $L_{\textit{out}}$. Specifically, $L_{\textit{in}}$ projects $h_{\textit{enc}}$ into a compatible dimension for $L_{\textit{LLM}}$, which then processes the projected representation. Subsequently, $L_{\textit{out}}$ converts the output of $L_{\textit{LLM}}$ back to the original hidden representation size of $h_{\textit{enc}}$. The final transformed feature is then fed into $\textit{Head}$ to generate the prediction $\hat{y}$. The overall process can be expressed as:

\begin{gather*}
h_{\textit{enc}} = \textit{Enc}(x), h_{\textit{LLM}} = L_{\textit{LLM}}(L_{\textit{in}}(h_{\textit{enc}})) \\
\hat{y} = \textit{Head}(L_{\textit{out}}(h_{\textit{LLM}}))
\end{gather*}

By introducing $L_{\textit{LLM}}$, PiFi enables SLMs to benefit from the rich knowledge encoded within LLMs, thereby improving performance on various downstream tasks.

\subsection{Plug-in of LLM Layer to Encoder-decoder LMs}

Encoder-decoder models are widely used for sequence-to-sequence tasks such as machine translation and text summarization. An encoder-decoder model consists of an encoder, $\textit{Enc}$, and a decoder, $\textit{Dec}$. The encoder $\textit{Enc}$ processes the input sequence $x$ to produce hidden representations $h_{\textit{enc}}$, which are then passed to $\textit{Dec}$ to generate the target sequence $\hat{y}$. This process can be represented as:

\begin{gather*}
h_{\textit{enc}} = \textit{Enc}(x), \hat{y}_{t} = \textit{Dec}(h_{\textit{enc}}, \hat{y}_{<t})
\end{gather*}

where $\hat{y}_{<t}$ denotes the tokens generated in previous time steps. The decoder $\textit{Dec}$ takes both $h{\textit{enc}}$ and $\hat{y}_{<t}$ as inputs to predict the next token $\hat{y}_{t}$.

In the PiFi framework, an LLM layer, $L_{\textit{LLM}}$, is inserted between $\textit{Enc}$ and $\textit{Dec}$. Similar to the procedure for encoder-based models, the hidden representation $h_{\textit{enc}}$ is first transformed using $L_{\textit{in}}$ to match the input size of $L_{\textit{LLM}}$. After processing by $L_{\textit{LLM}}$, the output is projected back to the original size using $L_{\textit{out}}$ before being fed into the decoder $\textit{Dec}$. This updated procedure for predicting $\hat{y}_{t}$ can be formulated as:

\begin{gather*}
h_{\textit{enc}} = \textit{Enc}(x), h_{\textit{LLM}} = L_{\textit{LLM}}(L_{\textit{in}}(h_{\textit{enc}})) \\
\hat{y}_{t} = \textit{Dec}(L_{\textit{out}}(h_{\textit{LLM}}), \hat{y}_{<t})
\end{gather*}

By incorporating $L_{\textit{LLM}}$ in this manner, PiFi utilizes the LLM’s knowledge to improve the performance of encoder-decoder models in generating high-quality target sequences.

\subsection{Fine-tuning of SLMs with Additional Layer}

During the fine-tuning stage, PiFi trains only the parameters of the original SLM, $L_{\textit{in}}$, $L_{\textit{out}}$, and the classification head. Importantly, the parameters of $L_{\textit{LLM}}$ are kept frozen and remain unchanged. This approach offers two key advantages: (1) it minimizes the number of additional parameters to be trained, and (2) it preserves the knowledge encoded in $L_{\textit{LLM}}$ during its pre-training stage. If we were to also update the parameters of $L_{\textit{LLM}}$, the model might suffer from catastrophic forgetting, a phenomenon where previously learned knowledge is lost when the model adapts to new data \cite{luo2023empirical, wang2023orthogonal}. By freezing the LLM layer, PiFi mitigates this risk and maintains the effectiveness of $L_{\textit{LLM}}$ throughout fine-tuning.

We evaluate the impact of freezing $L_{\textit{LLM}}$ through an ablation study, as presented in Appendix~\ref{app:ablation-full-ft}, demonstrating the effectiveness of this strategy in preventing catastrophic forgetting while ensuring optimal performance of the PiFi framework.

\section{Experiment}
\label{sec:experiment}
In this section, we present a comprehensive evaluation of our proposed PiFi framework through various experiments.

\subsection{Experimental Setup}
\label{sec:experiment-setup}

We conduct experiments using Llama-3.1-8B \cite{llama31english} as the default LLM to extract $L_{\textit{LLM}}$. Unless specified otherwise, the last layer of Llama-3.1-8B is integrated into a smaller LM as $L_{\textit{LLM}}$.

We performed our experiments on various tasks across NLU tasks and NLG tasks. For NLU tasks, we adopted text classification, natural language inference (NLI), and question answering (QA) tasks. Specifically, we used SST-2 \cite{socher2013recursive}, IMDB \cite{maas2011learning}, Tweet for sentiment classification and offensive language identification \cite{rosenthal2017semeval, zampieri2019semeval, barbieri2020tweeteval}, and CoLA \cite{warstadt2019neural} datasets for text classification tasks. For NLI tasks, we used MNLI \cite{williams2018broad} and SNLI \cite{bowman2015large}. For these two tasks, we measured the performance of the model through accuracy and F1-score. We used SQuAD v1.1 \cite{rajpurkar2016squad} for QA tasks, where the performance was measured by exact match and F1-score.

For NLG tasks, we evaluate each model through a machine translation task with Multi30k dataset \cite{elliott2016multi30k} and a text summarization task with CNN/DailyMail dataset \cite{nallapati2016abstractive}. For both tasks, we used BLEU \cite{papineni2002bleu}, ROUGE \cite{lin2004rouge}, METEOR \cite{banerjee2005meteor}, BERTScore \cite{zhang2020bertscore}, and BARTScore \cite{yuan2021bartscore} as a metric for measuring the performance of each model. All models were trained with five different random seeds, and we report the average performance for each experimental setup.

\begin{table*}[t]
\centering
\resizebox{0.9\textwidth}{!}{%
\begin{tabular}{l|ccccc|cc|c|c}
\Xhline{3\arrayrulewidth}
 &
  \multicolumn{5}{c|}{\textbf{Classification}} &
  \multicolumn{2}{c|}{\textbf{NLI}} &
  \textbf{QA} &
  \multirow{2}{*}{\textbf{Average}} \\ \cline{2-9}
 &
  \textbf{SST2} &
  \textbf{IMDB} &
  \textbf{\begin{tabular}[c]{@{}c@{}}Tweet\\ (Sentiment)\end{tabular}} &
  \textbf{\begin{tabular}[c]{@{}c@{}}Tweet\\ (Offensive)\end{tabular}} &
  \textbf{CoLA} &
  \textbf{MNLI} &
  \textbf{SNLI} &
  \textbf{SQuAD} &
   \\ \hline\hline 
\multicolumn{1}{l|}{BERT\textsubscript{\textit{base}}}&
  \begin{tabular}[c]{@{}c@{}}89.41\\ 0.8907\end{tabular} &
  \begin{tabular}[c]{@{}c@{}}85.1\\ 0.4733\end{tabular} &
  \begin{tabular}[c]{@{}c@{}}86.9\\ 0.862\end{tabular} &
  \begin{tabular}[c]{@{}c@{}}83.15\\ 0.7727\end{tabular} &
  \begin{tabular}[c]{@{}c@{}}80.10\\ 0.7398\end{tabular} &
  \begin{tabular}[c]{@{}c@{}}82.00\\ 0.8131\end{tabular} &
  \begin{tabular}[c]{@{}c@{}}89.10\\ 0.8892\end{tabular} &
  \begin{tabular}[c]{@{}c@{}}63.81\\ 0.7606\end{tabular} &
  \begin{tabular}[c]{@{}c@{}}82.45\\ 0.7752\end{tabular} \\
\rowcolor[HTML]{EFEFEF} 
\multicolumn{1}{l|}{\cellcolor[HTML]{EFEFEF}\makecell[{{l}}]{\textbf{+PiFi} \\ \textbf{(Llama-3.1-8B)}}} &
  \textbf{\begin{tabular}[c]{@{}c@{}}91.5\\ 0.9125\end{tabular}} &
  \textbf{\begin{tabular}[c]{@{}c@{}}87.09\\ 0.4800\end{tabular}} &
  \textbf{\begin{tabular}[c]{@{}c@{}}92.95\\ 0.9224\end{tabular}} &
  \textbf{\begin{tabular}[c]{@{}c@{}}86.03\\ 0.8026\end{tabular}} &
  \textbf{\begin{tabular}[c]{@{}c@{}}82.07\\ 0.7523\end{tabular}} &
  \textbf{\begin{tabular}[c]{@{}c@{}}82.74\\ 0.8185\end{tabular}} &
  \textbf{\begin{tabular}[c]{@{}c@{}}89.48\\ 0.8934\end{tabular}} &
  \textbf{\begin{tabular}[c]{@{}c@{}}66.17\\ 0.7809\end{tabular}} &
  \textbf{\begin{tabular}[c]{@{}c@{}}84.75\\ 0.7953\end{tabular}} \\ \hline
\multicolumn{1}{l|}{RoBERTa\textsubscript{\textit{base}}} &
  \begin{tabular}[c]{@{}c@{}}91.65\\ 0.9137\end{tabular} &
  \begin{tabular}[c]{@{}c@{}}87.36\\ 0.4890\end{tabular} &
  \begin{tabular}[c]{@{}c@{}}90.12\\ 0.894\end{tabular} &
  \begin{tabular}[c]{@{}c@{}}83.60\\ 0.7682\end{tabular} &
  \begin{tabular}[c]{@{}c@{}}80.55\\ 0.7192\end{tabular} &
  \begin{tabular}[c]{@{}c@{}}84.00\\ 0.8332\end{tabular} &
  \textbf{\begin{tabular}[c]{@{}c@{}}88.73\\ 0.8856\end{tabular}} &
  \begin{tabular}[c]{@{}c@{}}68.09\\ 0.8006\end{tabular} &
  \begin{tabular}[c]{@{}c@{}}84.32\\ 0.7879\end{tabular} \\
\rowcolor[HTML]{EFEFEF} 
\multicolumn{1}{l|}{\cellcolor[HTML]{EFEFEF}\makecell[{{l}}]{\textbf{+PiFi} \\ \textbf{(Llama-3.1-8B)}}} &
  \textbf{\begin{tabular}[c]{@{}c@{}}92.54\\ 0.9228\end{tabular}} &
  \textbf{\begin{tabular}[c]{@{}c@{}}88.68\\ 0.4878\end{tabular}} &
  \textbf{\begin{tabular}[c]{@{}c@{}}91.85\\ 0.9115\end{tabular}} &
  \textbf{\begin{tabular}[c]{@{}c@{}}85.35\\ 0.8083\end{tabular}} &
  \textbf{\begin{tabular}[c]{@{}c@{}}82.29\\ 0.7526\end{tabular}} &
  \textbf{\begin{tabular}[c]{@{}c@{}}84.25\\ 0.8346\end{tabular}} &
  \begin{tabular}[c]{@{}c@{}}88.49\\ 0.8837\end{tabular} &
  \textbf{\begin{tabular}[c]{@{}c@{}}68.97\\ 0.8089\end{tabular}} &
  \textbf{\begin{tabular}[c]{@{}c@{}}85.42\\ 0.7980\end{tabular}} \\ \hline
\multicolumn{1}{l|}{ELECTRA\textsubscript{\textit{base}}} &
  \begin{tabular}[c]{@{}c@{}}93.42\\ 0.9324\end{tabular} &
  \begin{tabular}[c]{@{}c@{}}88.31\\ 0.4974\end{tabular} &
  \begin{tabular}[c]{@{}c@{}}90.58\\ 0.8989\end{tabular} &
  \begin{tabular}[c]{@{}c@{}}83.52\\ 0.7751\end{tabular} &
  \begin{tabular}[c]{@{}c@{}}83.99\\ 0.7757\end{tabular} &
  \begin{tabular}[c]{@{}c@{}}85.41\\ 0.8472\end{tabular} &
  \begin{tabular}[c]{@{}c@{}}90.11\\ 0.8995\end{tabular} &
  \begin{tabular}[c]{@{}c@{}}44.44\\ 0.5706\end{tabular} &
  \begin{tabular}[c]{@{}c@{}}82.00\\ 0.7662\end{tabular} \\
\rowcolor[HTML]{EFEFEF} 
\multicolumn{1}{l|}{\cellcolor[HTML]{EFEFEF}\makecell[{{l}}]{\textbf{+PiFi} \\ \textbf{(Llama-3.1-8B)}}} &
  \textbf{\begin{tabular}[c]{@{}c@{}}94.13\\ 0.9393\end{tabular}} &
  \textbf{\begin{tabular}[c]{@{}c@{}}89.40\\ 0.4994\end{tabular}} &
  \textbf{\begin{tabular}[c]{@{}c@{}}0.9331\\ 0.9270\end{tabular}} &
  \textbf{\begin{tabular}[c]{@{}c@{}}84.99\\ 0.7848\end{tabular}} &
  \textbf{\begin{tabular}[c]{@{}c@{}}86.26\\ 0.8081\end{tabular}} &
  \textbf{\begin{tabular}[c]{@{}c@{}}86.47\\ 0.8618\end{tabular}} &
  \textbf{\begin{tabular}[c]{@{}c@{}}90.48\\ 0.9037\end{tabular}} &
  \textbf{\begin{tabular}[c]{@{}c@{}}67.99\\ 0.8045\end{tabular}} &
  \textbf{\begin{tabular}[c]{@{}c@{}}86.71\\ 0.8076\end{tabular}} \\ \hline
\multicolumn{1}{l|}{DeBERTa\textsubscript{\textit{base}}} &
  \begin{tabular}[c]{@{}c@{}}92.60\\ 0.9236\end{tabular} &
  \begin{tabular}[c]{@{}c@{}}87.98\\ 0.4882\end{tabular} &
  \begin{tabular}[c]{@{}c@{}}88.22\\ 0.8755\end{tabular} &
  \begin{tabular}[c]{@{}c@{}}82.67\\ 0.7729\end{tabular} &
  \begin{tabular}[c]{@{}c@{}}80.04\\ 0.7216\end{tabular} &
  \begin{tabular}[c]{@{}c@{}}83.72\\ 0.8312\end{tabular} &
  \begin{tabular}[c]{@{}c@{}}89.61\\ 0.8945\end{tabular} &
  \begin{tabular}[c]{@{}c@{}}67.87\\ 0.8094\end{tabular} &
  \begin{tabular}[c]{@{}c@{}}84.40\\ 0.7943\end{tabular} \\
\rowcolor[HTML]{EFEFEF} 
\multicolumn{1}{l|}{\cellcolor[HTML]{EFEFEF}\makecell[{{l}}]{\textbf{+PiFi} \\ \textbf{(Llama-3.1-8B)}}} &
  \textbf{\begin{tabular}[c]{@{}c@{}}93.04\\ 0.9283\end{tabular}} &
  \textbf{\begin{tabular}[c]{@{}c@{}}88.85\\ 0.4928\end{tabular}} &
  \textbf{\begin{tabular}[c]{@{}c@{}}92.17\\ 0.9152\end{tabular}} &
  \textbf{\begin{tabular}[c]{@{}c@{}}85.47\\ 0.8065\end{tabular}} &
  \textbf{\begin{tabular}[c]{@{}c@{}}80.87\\ 0.7347\end{tabular}} &
  \textbf{\begin{tabular}[c]{@{}c@{}}84.87\\ 0.8419\end{tabular}} &
  \textbf{\begin{tabular}[c]{@{}c@{}}90.62\\ 0.8982\end{tabular}} &
  \textbf{\begin{tabular}[c]{@{}c@{}}69.65\\ 0.8152\end{tabular}} &
  \textbf{\begin{tabular}[c]{@{}c@{}}85.63\\ 0.8037\end{tabular}} \\ \hline
\multicolumn{1}{l|}{DeBERTa-V3\textsubscript{\textit{base}}} &
  \begin{tabular}[c]{@{}c@{}}93.74\\ 0.9355\end{tabular} &
  \begin{tabular}[c]{@{}c@{}}89.45\\ 0.4951\end{tabular} &
  \begin{tabular}[c]{@{}c@{}}91.29\\ 0.9066\end{tabular} &
  \begin{tabular}[c]{@{}c@{}}83.60\\ 0.7915\end{tabular} &
  \begin{tabular}[c]{@{}c@{}}84.75\\ 0.8015\end{tabular} &
  \begin{tabular}[c]{@{}c@{}}87.52\\ 0.8687\end{tabular} &
  \begin{tabular}[c]{@{}c@{}}90.94\\ 0.9079\end{tabular} &
  \begin{tabular}[c]{@{}c@{}}69.40\\ 0.8249\end{tabular} &
  \begin{tabular}[c]{@{}c@{}}86.34\\ 0.8162\end{tabular} \\
\rowcolor[HTML]{EFEFEF} 
\multicolumn{1}{l|}{\cellcolor[HTML]{EFEFEF}\makecell[{{l}}]{\textbf{+PiFi} \\ \textbf{(Llama-3.1-8B)}}} &
  \textbf{\begin{tabular}[c]{@{}c@{}}95.01\\ 0.9481\end{tabular}} &
  \textbf{\begin{tabular}[c]{@{}c@{}}89.83\\ 0.5014\end{tabular}} &
  \textbf{\begin{tabular}[c]{@{}c@{}}93.80\\ 0.9325\end{tabular}} &
  \textbf{\begin{tabular}[c]{@{}c@{}}85.60\\ 0.8056\end{tabular}} &
  \textbf{\begin{tabular}[c]{@{}c@{}}86.07\\ 0.8118\end{tabular}} &
  \textbf{\begin{tabular}[c]{@{}c@{}}87.98\\ 0.8932\end{tabular}} &
  \textbf{\begin{tabular}[c]{@{}c@{}}91.05\\ 0.9095\end{tabular}} &
  \textbf{\begin{tabular}[c]{@{}c@{}}69.87\\ 0.8283\end{tabular}} &
  \textbf{\begin{tabular}[c]{@{}c@{}}87.40\\ 0.8287\end{tabular}} \\ \hline
\Xhline{3\arrayrulewidth}
\end{tabular}%
}
\caption{Experimental results of PiFi on various NLU tasks and datasets. We used the last layer from Llama-3.1-8B as $L_{\textit{LLM}}$ in this experiment. For classification and NLI tasks, we report accuracy and F1-score in the upper row and lower row of each cell. For the QA task, we report the exact match and F1-score in the upper row and lower row of each cell.}
\label{tab:tab1-nlu}
\end{table*}
% \usepackage{graphicx}
\begin{table*}[t]
\renewcommand{\arraystretch}{1.5} % Adjust this number to change row height
\centering
\resizebox{0.9\textwidth}{!}{%
\begin{tabular}{l|ccccc|ccccc}
\Xhline{3\arrayrulewidth}
\rowcolor[HTML]{FFFFFF} 
\multicolumn{1}{c|}{\cellcolor[HTML]{FFFFFF}\textbf{}} &
  \multicolumn{5}{c|}{\cellcolor[HTML]{FFFFFF}\textbf{Multi30K (Translation)}} &
  \multicolumn{5}{c}{\cellcolor[HTML]{FFFFFF}\textbf{CNN/Daily Mail (Summarization)}} \\ \cline{2-11}  
\rowcolor[HTML]{FFFFFF} 
\multicolumn{1}{c|}{\cellcolor[HTML]{FFFFFF}\textbf{}} &
  \textbf{BLEU} &
  \textbf{ROUGE} &
  \textbf{METEOR} &
  \textbf{BERTS.} &
  \textbf{BARTS.} &
  \textbf{BLEU} &
  \textbf{ROUGE} &
  \textbf{METEOR} &
  \textbf{BERTS.} &
  \textbf{BARTS.} \\ \hline\hline
\rowcolor[HTML]{FFFFFF} 
\multicolumn{1}{l|}{T5\textsubscript{\textit{base}}}&
  0.5301 &
  0.6195 &
  \textbf{0.3605} &
  0.8724 &
  \textbf{-4.634} &
  0.2175 &
  0.2323 &
  0.1731 &
  0.7409 &
  -5.784 \\
\rowcolor[HTML]{EFEFEF} 
\multicolumn{1}{l|}{\cellcolor[HTML]{EFEFEF}\makecell[{{l}}]{\textbf{+ PiFi} \\ \textbf{(Llama-3.1-8B)}}} &
  \textbf{0.5413} &
  \textbf{0.6536} &
  0.3534 &
  \textbf{0.8978} &
  -4.669 &
  \textbf{0.2242} &
  \textbf{0.2357} &
  \textbf{0.1752} &
  \textbf{0.7412} &
  \textbf{-5.777} \\ \hline
\rowcolor[HTML]{FFFFFF} 
\multicolumn{1}{l|}{BART\textsubscript{\textit{base}}} &
  0.4580 &
  0.5864 &
  0.3331 &
  \textbf{0.8635} &
  \textbf{-4.513} &
  0.2270 &
  0.2348 &
  {\color[HTML]{363A3D} 0.1782} &
  0.7424 &
  -5.665 \\
\rowcolor[HTML]{EFEFEF} 
\multicolumn{1}{l|}{\cellcolor[HTML]{EFEFEF}\makecell[{{l}}]{\textbf{+ PiFi} \\ \textbf{(Llama-3.1-8B)}}} &
  {\color[HTML]{333333} \textbf{0.4695}} &
  {\color[HTML]{333333} \textbf{0.5908}} &
  {\color[HTML]{333333} \textbf{0.3364}} &
  {\color[HTML]{333333} 0.8617} &
  {\color[HTML]{333333} -4.515} &
  {\color[HTML]{333333} \textbf{0.2331}} &
  {\color[HTML]{333333} \textbf{0.2355}} &
  {\color[HTML]{333333} \textbf{0.1799}} &
  {\color[HTML]{333333} \textbf{0.7425}} &
  {\color[HTML]{333333} \textbf{-5.652}} \\
\Xhline{3\arrayrulewidth}

\end{tabular}%
}
\caption{Experimental results of PiFi on machine translation and text summarization tasks. We used the last layer from Llama-3.1-8B as $L_{\textit{LLM}}$ in this experiment.}
\label{tab:tab2-nlg}
\end{table*}

\subsection{Performance Improvement of PiFi on NLU Tasks}
\label{sec:experiment-nlu}

We assess PiFi’s effectiveness on NLU tasks using several SLMs: BERT \cite{devlin2019bert}, RoBERTa \cite{liu2019roberta}, ELECTRA \cite{clark2020electra}, DeBERTa \cite{he2021deberta}, and DeBERTaV3 \cite{he2023debertav}.

Table~\ref{tab:tab1-nlu} shows the results, where PiFi consistently outperforms vanilla fine-tuned SLMs across all datasets. For example, integrating PiFi into BERT resulted in a 2.3\%p increase in average accuracy compared to standard fine-tuning. Similar gains are observed for other models, demonstrating PiFi’s compatibility and effectiveness across a diverse set of NLU tasks and architectures.

\subsection{Performance Improvement of PiFi on NLG Tasks}
\label{sec:experiment-nlg}

To validate PiFi on NLG tasks, we employed encoder-decoder models such as T5\textsubscript{\textit{base}} \cite{raffel2020exploring} and BART\textsubscript{\textit{base}} \cite{lewis2020bart}. These models were evaluated on machine translation and text summarization tasks, thereby validating the advantage of PiFi on NLG tasks beyond NLU tasks.

The experimental results on two NLG tasks are displayed in Table~\ref{tab:tab2-nlg}. We found that the SLM trained with PiFi exhibits higher performance compared to vanilla fine-tuned SLM in most cases. Particularly, for the text summarization task, both models trained with PiFi achieved higher scores across all evaluation metrics, indicating that PiFi effectively transfers knowledge from the LLM to SLMs, thus enhancing linguistic capabilities in NLG tasks.

% \usepackage{multirow}
% \usepackage{graphicx}
\begin{table}[t]
\centering
\resizebox{\columnwidth}{!}{%
\begin{tabular}{c|l|ccc}
\Xhline{5\arrayrulewidth}
\multicolumn{1}{l|}{} &
   &
  \multicolumn{3}{c}{\textbf{Test Dataset}} \\ \cline{3-5} 
\begin{tabular}[c]{@{}c@{}}\textbf{Train}\\ \textbf{Dataset}\end{tabular} &
   &
  \textbf{IMDB} &
  \begin{tabular}[c]{@{}c@{}}\textbf{Tweet}\\ \textbf{(Sentiment)}\end{tabular} &
  \textbf{CR} \\ \hline\hline
 &
  BERT\textsubscript{\textit{base}} &
  \begin{tabular}[c]{@{}c@{}}85.1\\ 0.4773\end{tabular} &
  \begin{tabular}[c]{@{}c@{}}70.40\\ 0.6918\end{tabular} &
  \begin{tabular}[c]{@{}c@{}}74.56\\ 0.7301\end{tabular} \\
IMDB &
  \cellcolor[HTML]{EFEFEF}\textbf{\begin{tabular}[c]{@{}l@{}}+PiFi\\ (Llam-3-8B)\end{tabular}} &
  \cellcolor[HTML]{EFEFEF}\textbf{\begin{tabular}[c]{@{}c@{}}87.09\\ 0.4800\end{tabular}} &
  \cellcolor[HTML]{EFEFEF}\textbf{\begin{tabular}[c]{@{}c@{}}83.68\\ 0.8176\end{tabular}} &
  \cellcolor[HTML]{EFEFEF}\textbf{\begin{tabular}[c]{@{}c@{}}79.86\\ 0.788\end{tabular}} \\
\multicolumn{1}{l|}{} &
  BERT\textsubscript{\textit{large}} &
  \begin{tabular}[c]{@{}c@{}}86.88\\ 0.4836\end{tabular} &
  \begin{tabular}[c]{@{}c@{}}77.39\\ 0.7647\end{tabular} &
  \begin{tabular}[c]{@{}c@{}}76.10\\ 0.7419\end{tabular} \\ \hline
 &
  BERT\textsubscript{\textit{base}} &
  \begin{tabular}[c]{@{}c@{}}74.70\\ 0.4306\end{tabular} &
  \begin{tabular}[c]{@{}c@{}}86.90\\ 0.862\end{tabular} &
  \begin{tabular}[c]{@{}c@{}}85.46\\ 0.8357\end{tabular} \\
\begin{tabular}[c]{@{}c@{}}Tweet\\ (Sentiment)\end{tabular} &
  \cellcolor[HTML]{EFEFEF}\textbf{\begin{tabular}[c]{@{}l@{}}+PiFi\\ (Llam-3-8B)\end{tabular}} &
  \cellcolor[HTML]{EFEFEF}{\color[HTML]{363A3D} \textbf{\begin{tabular}[c]{@{}c@{}}77.28\\ 0.4351\end{tabular}}} &
  \cellcolor[HTML]{EFEFEF}\textbf{\begin{tabular}[c]{@{}c@{}}92.95\\ 0.9224\end{tabular}} &
  \cellcolor[HTML]{EFEFEF}\textbf{\begin{tabular}[c]{@{}c@{}}87.25\\ 0.8596\end{tabular}} \\
\multicolumn{1}{l|}{} &
  BERT\textsubscript{\textit{large}} &
  {\color[HTML]{363A3D} \begin{tabular}[c]{@{}c@{}}75.91\\ 0.4331\end{tabular}} &
  \begin{tabular}[c]{@{}c@{}}89.26\\ 0.8853\end{tabular} &
  \begin{tabular}[c]{@{}c@{}}86.50\\ 0.8526\end{tabular} \\ \hline
 &
  BERT\textsubscript{\textit{base}} &
  \begin{tabular}[c]{@{}c@{}}75.72\\ 0.4301\end{tabular} &
  \begin{tabular}[c]{@{}c@{}}82.52\\ 0.8161\end{tabular} &
  \begin{tabular}[c]{@{}c@{}}89.60\\ 0.8857\end{tabular} \\
CR &
  \cellcolor[HTML]{EFEFEF}\textbf{\begin{tabular}[c]{@{}l@{}}+PiFi\\ (Llam-3-8B)\end{tabular}} &
  \cellcolor[HTML]{EFEFEF}\textbf{\begin{tabular}[c]{@{}c@{}}77.49\\ 0.4362\end{tabular}} &
  \cellcolor[HTML]{EFEFEF}\textbf{\begin{tabular}[c]{@{}c@{}}84.80\\ 0.8365\end{tabular}} &
  \cellcolor[HTML]{EFEFEF}\textbf{\begin{tabular}[c]{@{}c@{}}90.90\\ 0.9015\end{tabular}} \\
\multicolumn{1}{l|}{} &
  BERT\textsubscript{\textit{large}} &
  \begin{tabular}[c]{@{}c@{}}76.77\\ 0.4341\end{tabular} &
  \begin{tabular}[c]{@{}c@{}}83.90\\ 0.8284\end{tabular} &
  \begin{tabular}[c]{@{}c@{}}90.64\\ 0.8989\end{tabular} \\
\Xhline{5\arrayrulewidth}
\end{tabular}%
}
\caption{Experimental results of PiFi under domain shift. We report accuracy and F1-score in the upper row and lower row of each cell.}
\label{tab:tab3-domain}
\end{table}
% \usepackage{graphicx}
\begin{table}[t]
\centering
\resizebox{0.9\columnwidth}{!}{%
\begin{tabular}{l|cc}
\Xhline{3\arrayrulewidth}

\hline
\multicolumn{1}{c|}{} &
  \textbf{\begin{tabular}[c]{@{}c@{}}NSMC\\ (Korean)\end{tabular}} &
  \textbf{\begin{tabular}[c]{@{}c@{}}Filmstarts\\ (German)\end{tabular}} \\ \hline\hline
\multicolumn{1}{l|}{{mBERT\textsubscript{\textit{base}}}} &
  \begin{tabular}[c]{@{}c@{}}83.62\\ 0.8318\end{tabular} &
  \begin{tabular}[c]{@{}c@{}}86.77\\ 0.8279\end{tabular} \\ \hline
\begin{tabular}[c]{@{}l@{}}\textbf{+ PiFi}\\ \textbf{(Llama-3-8B English)}\end{tabular} &
  \begin{tabular}[c]{@{}c@{}}84.04\\ 0.8362\end{tabular} &
  \begin{tabular}[c]{@{}c@{}}87.92\\ 0.8362\end{tabular} \\ \hline
\begin{tabular}[c]{@{}l@{}}\textbf{+ PiFi}\\ \textbf{(Llama-3-8B Korean})\end{tabular} &
  \textbf{\begin{tabular}[c]{@{}c@{}}85.61\\ 0.8522\end{tabular}} &
  \begin{tabular}[c]{@{}c@{}}87.09\\ 0.8337\end{tabular} \\ \hline
\begin{tabular}[c]{@{}l@{}}\textbf{+ PiFi}\\ \textbf{(Llama-3-8B German)}\end{tabular} &
  \begin{tabular}[c]{@{}c@{}}83.85\\ 0.8341\end{tabular} &
  \textbf{\begin{tabular}[c]{@{}c@{}}88.11\\ 0.8411\end{tabular}} \\ 

\Xhline{3\arrayrulewidth}
\end{tabular}%
}
\caption{Experimental results on Korean and German text classification datasets. For this experiment, we used Llama-3 model trained in English, Korean, and German. We report accuracy and F1-score in the upper row and lower row of each cell.}
\label{tab:tab4-linguistic}
\end{table}

\subsection{Generalizability of PiFi under Domain Shift}
\label{sec:experiment-domain}

Since LLMs are pre-trained on much larger and more diverse corpora compared to SLMs, integrating LLMs into SLMs through PiFi is expected to improve the generalizability of the model, particularly in unseen domains. To test this hypothesis, we conducted an experiment comparing the performance of the PiFi model against a vanilla fine-tuned model under conditions of domain shift. For this experiment, we used three text classification datasets. Specifically, we used IMDB, CR \cite{ding2008holistic}, and Tweet for sentiment classification. While all three datasets involve sentiment classification, they represent different domains such as movie review, electronics product review, and tweet messages. We trained the models on each dataset and evaluated their performance across all three datasets.

The results, shown in Table~\ref{tab:tab3-domain}, indicate that PiFi consistently outperforms vanilla fine-tuning across all domain shifts. Notably, the PiFi model trained on IMDB showed significant performance improvements when tested on the Tweet and CR datasets, with gains of 13.28\%p and 5.3\%p, respectively, over the vanilla fine-tuned model. To verify that this improved generalization is due to leveraging LLM layer knowledge rather than merely a simple increase in parameters, we compared PiFi’s fine-tuning performance against BERT\textsubscript{\textit{large}}, which is slightly larger than the PiFi model. The results showed that PiFi outperformed BERT\textsubscript{\textit{large}}, confirming that incorporating the LLM layer into an SLM via PiFi allows the model to effectively leverage the extensive knowledge stored in the LLM. This enhancement offers benefits beyond parameter scaling, thereby improving its ability to generalize to unseen domains.

\subsection{Transferring of Linguistic Ability of LLMs to SLMs}
\label{sec:experiment-linguistic}

In this section, we examine how the primary language of large language models (LLMs) influences the performance of the PiFi model when applied to smaller language models (SLMs). Specifically, we explore this effect by training an mBERT model \cite{pires2019multilingual} on two distinct datasets: NSMC \cite{park2016nsmc}, a Korean sentiment classification dataset for movie reviews, and Filmstarts \cite{guhr2020training}, a German movie review sentiment dataset. For our experiments, we use Llama-3-8B\footnote{Note that Llama 3, not Llama 3.1, was used in this experiment to ensure a fair comparison across English, German, and Korean models.} along with Llama-3-8B variants further fine-tuned for Korean and German \cite{llama3openko, llama3german} to extract $L_{\textit{LLM}}$, which is then integrated into the mBERT model.

The results of our experiments are presented in Table~\ref{tab:tab4-linguistic}. We find that while PiFi demonstrates performance improvements on non-English datasets using the default English Llama-3 model, utilizing language-specific Llama-3 variants (e.g., Korean or German) leads to further performance gains for their respective languages. For example, the PiFi model, when combined with the Korean Llama-3 model, achieved an additional 1.57\%p improvement over the English Llama-3 model on the Korean dataset. Conversely, when a PiFi model is trained on a different language than the target downstream task, the performance gains diminish. For instance, training PiFi on the Filmstarts (German) dataset with the Korean Llama-3 model resulted in a performance increase that was 0.83\%p lower than that obtained with the English Llama-3 model and 1.02\%p lower than that with the German Llama-3 model.

These findings indicate that the PiFi model benefits primarily from effectively leveraging knowledge from the LLM through a single layer, rather than simply from an increase in the number of model parameters. In conclusion, aligning the language of the downstream task with the language of the LLM used for $L_{\textit{LLM}}$ extraction is critical for maximizing the effectiveness of PiFi, as this alignment ensures optimal utilization of the LLM's linguistic capabilities.

% \usepackage{graphicx}
\begin{table}[t]
\centering
\resizebox{\columnwidth}{!}{%
\begin{tabular}{l|ccccc}
\Xhline{5\arrayrulewidth}

 &
  {\color[HTML]{0E0E0E} SST-2} &
  {\color[HTML]{0E0E0E} IMDB} &
  {\color[HTML]{0E0E0E} \begin{tabular}[c]{@{}c@{}}Tweet\\ (Sentiment)\end{tabular}} &
  {\color[HTML]{0E0E0E} \begin{tabular}[c]{@{}c@{}}Tweet\\ (Offensive)\end{tabular}} &
  {\color[HTML]{0E0E0E} CoLA} \\ \hline\hline
BERT\textsubscript{\textit{base}} &
  {\color[HTML]{0E0E0E} \begin{tabular}[c]{@{}c@{}}89.41\\ 0.8907\end{tabular}} &
  {\color[HTML]{0E0E0E} \begin{tabular}[c]{@{}c@{}}85.10\\ 0.4733\end{tabular}} &
  {\color[HTML]{0E0E0E} \begin{tabular}[c]{@{}c@{}}86.90\\ 0.862\end{tabular}} &
  {\color[HTML]{0E0E0E} \begin{tabular}[c]{@{}c@{}}83.15\\ 0.7727\end{tabular}} &
  {\color[HTML]{0E0E0E} \begin{tabular}[c]{@{}c@{}}80.10\\ 0.7398\end{tabular}} \\ \hline
{\color[HTML]{0E0E0E} \begin{tabular}[c]{@{}l@{}}+ PiFi-Random\\ (Llama-3.1-8B)\end{tabular}} &
  \begin{tabular}[c]{@{}c@{}}90.28\\ 0.8997\end{tabular} &
  \begin{tabular}[c]{@{}c@{}}86.43\\ 0.4879\end{tabular} &
  {\color[HTML]{0E0E0E} \begin{tabular}[c]{@{}c@{}}89.49\\ 0.9095\end{tabular}} &
  {\color[HTML]{0E0E0E} \begin{tabular}[c]{@{}c@{}}83.50\\ 0.7564\end{tabular}} &
  {\color[HTML]{0E0E0E} \begin{tabular}[c]{@{}c@{}}80.01\\ 0.7046\end{tabular}} \\ \hline
{\color[HTML]{0E0E0E} \begin{tabular}[c]{@{}l@{}}+ PiFi-Random-Full\\ (Llama-3.1-8B)\end{tabular}} &
  {\color[HTML]{0E0E0E} \begin{tabular}[c]{@{}c@{}}90.50\\ 0.9017\end{tabular}} &
  {\color[HTML]{0E0E0E} \begin{tabular}[c]{@{}c@{}}86.46\\ 0.4767\end{tabular}} &
  {\color[HTML]{0E0E0E} \begin{tabular}[c]{@{}c@{}}90.95\\ 0.9024\end{tabular}} &
  {\color[HTML]{0E0E0E} \begin{tabular}[c]{@{}c@{}}83.95\\ 0.7812\end{tabular}} &
  \begin{tabular}[c]{@{}c@{}}80.39\\ 0.7376\end{tabular} \\ \hline
{\color[HTML]{0E0E0E} \textbf{\begin{tabular}[c]{@{}l@{}}+ PiFi\\ (Llama-3.1-8B)\end{tabular}}} &
  \textbf{\begin{tabular}[c]{@{}c@{}}91.50\\ 0.9125\end{tabular}} &
  \textbf{\begin{tabular}[c]{@{}c@{}}87.09\\ 0.48\end{tabular}} &
  \textbf{\begin{tabular}[c]{@{}c@{}}92.95\\ 0.9224\end{tabular}} &
  \textbf{\begin{tabular}[c]{@{}c@{}}86.03\\ 0.8026\end{tabular}} &
  {\color[HTML]{0E0E0E} \textbf{\begin{tabular}[c]{@{}c@{}}82.07\\ 0.7523\end{tabular}}}\\
\Xhline{5\arrayrulewidth}
  
\end{tabular}%
}
\caption{Experimental results of PiFi under different configurations of $L_{LLM}$ initialization and fine-tuning. For this experiment, we compared PiFi-Random (frozen randomly initialized $L_{LLM}$), PiFi-Random-Full (fine-tuned randomly initialized $L_{LLM}$), and PiFi (pre-trained $L_{LLM}$). We report accuracy and F1-score in the upper row and lower row of each cell.
}
\label{tab:tab5-random}
\end{table}
% \usepackage{graphicx}
\begin{table}[t]
\centering
\resizebox{\columnwidth}{!}{%
\begin{tabular}{l|c|ccccc}
\Xhline{5\arrayrulewidth}

 &
  \begin{tabular}[c]{@{}c@{}}Params\\ (M)\end{tabular} &
  SST-2 &
  IMDB &
  \begin{tabular}[c]{@{}c@{}}Tweet\\ (Sentiment)\end{tabular} &
  \begin{tabular}[c]{@{}c@{}}Tweet\\ (Offensive)\end{tabular} &
  CoLA \\ \hline\hline
BERT\textsubscript{\textit{base}} &
  110 &
  \begin{tabular}[c]{@{}c@{}}89.41\\ 0.8907\end{tabular} &
  \begin{tabular}[c]{@{}c@{}}85.10\\ 0.4733\end{tabular} &
  \begin{tabular}[c]{@{}c@{}}86.90\\ 0.8620\end{tabular} &
  \begin{tabular}[c]{@{}c@{}}83.15\\ 0.7727\end{tabular} &
  \begin{tabular}[c]{@{}c@{}}80.10\\ 0.7398\end{tabular} \\ \hline
\textbf{\begin{tabular}[c]{@{}l@{}}+PiFi\\ (Llama-3.1-8B)\end{tabular}}  &
  334 &
  \textbf{\begin{tabular}[c]{@{}c@{}}91.50\\ 0.9125\end{tabular}} &
  \textbf{\begin{tabular}[c]{@{}c@{}}87.09\\ 0.48\end{tabular}} &
  \textbf{\begin{tabular}[c]{@{}c@{}}92.95\\ 0.9224\end{tabular}} &
  \textbf{\begin{tabular}[c]{@{}c@{}}86.03\\ 0.8026\end{tabular}} &
  \textbf{\begin{tabular}[c]{@{}c@{}}82.07\\ 0.7523\end{tabular}} \\ \hline
BERT\textsubscript{\textit{large}} &
  336 &
  \begin{tabular}[c]{@{}c@{}}90.73\\ 0.9040\end{tabular} &
  \begin{tabular}[c]{@{}c@{}}86.88\\ 0.4836\end{tabular} &
  \begin{tabular}[c]{@{}c@{}}89.26\\ 0.8853\end{tabular} &
  \begin{tabular}[c]{@{}c@{}}83.95\\ 0.7887\end{tabular} &
  \begin{tabular}[c]{@{}c@{}}80.96\\ 0.7327\end{tabular}\\
\Xhline{5\arrayrulewidth}
  
\end{tabular}%
}
\caption{Experimental results comparing the performance of PiFi and BERT\textsubscript{\textit{large}} with similar parameter counts. We report accuracy and F1-score in the upper row and lower row of each cell.}
\label{tab:tab6-large}
\end{table}
% \usepackage{graphicx}
\begin{table}[t]
\centering
\resizebox{\columnwidth}{!}{%
\begin{tabular}{l|ccccc}
\Xhline{3\arrayrulewidth}

 &
  \textbf{SST-2} &
  \textbf{IMDB} &
  \begin{tabular}[c]{@{}c@{}}\textbf{Tweet}\\ \textbf{(Sentiment)}\end{tabular} &
  \begin{tabular}[c]{@{}c@{}}\textbf{Tweet}\\ \textbf{(Offensive)}\end{tabular} &
  \textbf{CoLA} \\ \hline\hline

BERT\textsubscript{\textit{base}} &
  \begin{tabular}[c]{@{}c@{}}89.41\\ 0.8907\end{tabular} &
  \begin{tabular}[c]{@{}c@{}}85.10\\ 0.4733\end{tabular} &
  \begin{tabular}[c]{@{}c@{}}86.90\\ 0.862\end{tabular} &
  \begin{tabular}[c]{@{}c@{}}83.15\\ 0.7727\end{tabular} &
  \begin{tabular}[c]{@{}c@{}}80.10\\ 0.7398\end{tabular} \\\hline
\begin{tabular}[c]{@{}l@{}}\textbf{+ PiFi}\\ \textbf{(Llama-3.1-8B)}\end{tabular} &
  \textbf{\begin{tabular}[c]{@{}c@{}}91.50\\ 0.9125\end{tabular}} &
  \textbf{\begin{tabular}[c]{@{}c@{}}87.09\\ 0.48\end{tabular}} &
  \textbf{\begin{tabular}[c]{@{}c@{}}92.95\\ 0.9224\end{tabular}} &
  \textbf{\begin{tabular}[c]{@{}c@{}}86.03\\ 0.8026\end{tabular}} &
  \textbf{\begin{tabular}[c]{@{}c@{}}82.07\\ 0.7523\end{tabular}} \\ \hline
  
\begin{tabular}[c]{@{}l@{}}\textbf{+ PiFi}\\ \textbf{(Mistral-7B-v0.1)}\end{tabular} &
  \begin{tabular}[c]{@{}c@{}}90.89\\ 0.9061\end{tabular} &
  \begin{tabular}[c]{@{}c@{}}86.47\\ 0.4794\end{tabular} &
  {\color[HTML]{363A3D} \begin{tabular}[c]{@{}c@{}}90.12\\ 0.8943\end{tabular}} &
  \begin{tabular}[c]{@{}c@{}}84.99\\ 0.8055\end{tabular} &
  \begin{tabular}[c]{@{}c@{}}81.65\\ 0.7324\end{tabular} \\\hline
\begin{tabular}[c]{@{}l@{}}\textbf{+ PiFi}\\ \textbf{(Mistral-7B-v0.3)}\end{tabular} &
  \begin{tabular}[c]{@{}c@{}}91.65\\ 0.9136\end{tabular} &
  \begin{tabular}[c]{@{}c@{}}87.22\\ 0.4797\end{tabular} &
  \begin{tabular}[c]{@{}c@{}}92.57\\ 0.9177\end{tabular} &
  \begin{tabular}[c]{@{}c@{}}85.33\\ 0.802\end{tabular} &
  \begin{tabular}[c]{@{}c@{}}81.72\\ 0.7502\end{tabular} \\\hline
\begin{tabular}[c]{@{}l@{}}\textbf{+ PiFi}\\ \textbf{(Qwen2 -7B)}\end{tabular} &
  \begin{tabular}[c]{@{}c@{}}91.17\\ 0.9092\end{tabular} &
  \begin{tabular}[c]{@{}c@{}}86.68\\ 0.4774\end{tabular} &
  \begin{tabular}[c]{@{}c@{}}92.48\\ 0.917\end{tabular} &
  \begin{tabular}[c]{@{}c@{}}85.70\\ 0.809\end{tabular} &
  \begin{tabular}[c]{@{}c@{}}81.62\\ 0.7475\end{tabular} \\\hline
\begin{tabular}[c]{@{}l@{}}\textbf{+ PiFi}\\ \textbf{(Gemma-2-9B)}\end{tabular} &
  {\color[HTML]{363A3D} \begin{tabular}[c]{@{}c@{}}91.39\\ 0.9111\end{tabular}} &
  \begin{tabular}[c]{@{}c@{}}87.1\\ 0.4849\end{tabular} &
  \begin{tabular}[c]{@{}c@{}}92.34\\ 0.9163\end{tabular} &
  \begin{tabular}[c]{@{}c@{}}84.29\\ 0.7866\end{tabular} &
  \begin{tabular}[c]{@{}c@{}}80.74\\ 0.7598\end{tabular} \\\hline
\begin{tabular}[c]{@{}l@{}}\textbf{+ PiFi}\\ \textbf{(Falcon-7B)}\end{tabular} &
  \begin{tabular}[c]{@{}c@{}}91.44\\ 0.9115\end{tabular} &
  \begin{tabular}[c]{@{}c@{}}86.63\\ 0.4779\end{tabular} &
  \begin{tabular}[c]{@{}c@{}}92.51\\ 0.9178\end{tabular} &
  \begin{tabular}[c]{@{}c@{}}84.85\\ 0.7907\end{tabular} &
  \begin{tabular}[c]{@{}c@{}}80.99\\ 0.7518\end{tabular} \\
\Xhline{3\arrayrulewidth}
\end{tabular}%
}
\caption{Experimental results of PiFi across various LLMs, such as Llama-3.1-8B, Mistral-7B-v0.1 and v0.3, Qwen-7B, Gemma-2-9B, and Falcon-7B. We extracted the last layer from each LLM as $L_{\textit{LLM}}$ and compared their performance. We report accuracy and F1-score in the upper row and lower row of each cell.}
\label{tab:tab5-llm}
\end{table}

\subsection{Probing the Role of Intrinsic Knowledge in LLM Layer}
\label{sec:experiment-random}

To verify that the performance improvement achieved by plugging in $L_{\textit{LLM}}$ is not simply due to an increase in the number of parameters but rather a result of leveraging the inherent knowledge embedded in the LLM, we designed a series of experiments.

First, Table~\ref{tab:tab5-random} shows the results of experiments conducted with randomly initialized $L_{\textit{LLM}}$ under two configurations. PiFi-Random refers to a model where the randomly initialized $L_{\textit{LLM}}$ is frozen during fine-tuning, while PiFi-Random-Full is a model where the entire network, including the randomly initialized $L_{\textit{LLM}}$, is fine-tuned. Although PiFi-Random and PiFi-Random-Full exhibited slight improvements over BERT\textsubscript{\textit{base}}, their performance fell short of PiFi, which utilizes the pre-trained weights of the $L_{\textit{LLM}}$. These results demonstrate that the benefits of simply increasing the number of parameters are limited.

In addition to the comparison with randomly initialized baselines, Table~\ref{tab:tab6-large} compares the performance of PiFi with BERT\textsubscript{\textit{large}}, which has a similar number of parameters. BERT\textsubscript{\textit{large}} has approximately 336M parameters, while PiFi applied to BERT\textsubscript{\textit{base}} has a comparable parameter count of approximately 334M. Despite this similarity in parameters and native architecture of BERT\textsubscript{\textit{large}}, BERT\textsubscript{\textit{base}} with PiFi outperformed BERT\textsubscript{\textit{large}} in terms of performance.

This experimental finding suggests that the performance improvement of PiFi is not merely an outcome of increasing the number of parameters but is primarily due to effectively utilizing the pre-trained knowledge embedded in the LLM.

% \usepackage{graphicx}
\begin{table}[t]
\centering
\resizebox{\columnwidth}{!}{%
\begin{tabular}{l|ccccc}
\Xhline{3\arrayrulewidth}

 &
  {\color[HTML]{0E0E0E} \textbf{SST-2}} &
  {\color[HTML]{0E0E0E} \textbf{IMDB}} &
  {\color[HTML]{0E0E0E} \begin{tabular}[c]{@{}c@{}}\textbf{Tweet}\\ \textbf{(Sentiment)}\end{tabular}} &
  {\color[HTML]{0E0E0E} \begin{tabular}[c]{@{}c@{}}\textbf{Tweet}\\ \textbf{(Offensive)}\end{tabular}} &
  {\color[HTML]{0E0E0E} \textbf{CoLA}} \\ \hline\hline
BERT\textsubscript{\textit{base}} &
  {\color[HTML]{0E0E0E} \begin{tabular}[c]{@{}c@{}}89.41\\ 0.8907\end{tabular}} &
  {\color[HTML]{0E0E0E} \begin{tabular}[c]{@{}c@{}}85.1\\ 0.4733\end{tabular}} &
  {\color[HTML]{0E0E0E} \begin{tabular}[c]{@{}c@{}}86.90\\ 0.862\end{tabular}} &
  {\color[HTML]{0E0E0E} \begin{tabular}[c]{@{}c@{}}83.15\\ 0.7727\end{tabular}} &
  {\color[HTML]{0E0E0E} \begin{tabular}[c]{@{}c@{}}80.10\\ 0.7398\end{tabular}} \\ \hline
{\color[HTML]{0E0E0E} \begin{tabular}[c]{@{}l@{}}\textbf{+ PiFi}\\ \textbf{(Qwen2-0.5B)}\end{tabular}} &
  {\color[HTML]{0E0E0E} \begin{tabular}[c]{@{}c@{}}90.13\\ 0.8977\end{tabular}} &
  {\color[HTML]{0E0E0E} \begin{tabular}[c]{@{}c@{}}86.09\\ 0.4843\end{tabular}} &
  {\color[HTML]{0E0E0E} \begin{tabular}[c]{@{}c@{}}89.67\\ 0.8899\end{tabular}} &
  {\color[HTML]{0E0E0E} \begin{tabular}[c]{@{}c@{}}84.52\\ 0.7959\end{tabular}} &
  {\color[HTML]{0E0E0E} \begin{tabular}[c]{@{}c@{}}81.15\\ 0.7453\end{tabular}} \\ \hline
{\color[HTML]{0E0E0E} \begin{tabular}[c]{@{}l@{}}\textbf{+ PiFi}\\ \textbf{(Qwen2-1.5B)}\end{tabular}} &
  {\color[HTML]{0E0E0E} \begin{tabular}[c]{@{}c@{}}91.06\\ 0.9079\end{tabular}} &
  {\color[HTML]{0E0E0E} \begin{tabular}[c]{@{}c@{}}86.29\\ 0.4668\end{tabular}} &
  {\color[HTML]{0E0E0E} \begin{tabular}[c]{@{}c@{}}92.35\\ 0.9164\end{tabular}} &
  {\color[HTML]{0E0E0E} \begin{tabular}[c]{@{}c@{}}85.91\\ 0.8096\end{tabular}} &
  \begin{tabular}[c]{@{}c@{}}81.54\\ 0.7453\end{tabular} \\ \hline
{\color[HTML]{0E0E0E} \begin{tabular}[c]{@{}l@{}}\textbf{+ PiFi}\\ \textbf{(Qwen2-7B)}\end{tabular}} &
  \textbf{\begin{tabular}[c]{@{}c@{}}91.17\\ 0.9092\end{tabular}} &
  \textbf{\begin{tabular}[c]{@{}c@{}}86.68\\ 0.4774\end{tabular}} &
  \textbf{\begin{tabular}[c]{@{}c@{}}92.48\\ 0.917\end{tabular}} &
  \textbf{\begin{tabular}[c]{@{}c@{}}86.04\\ 0.8068\end{tabular}} &
  {\color[HTML]{0E0E0E} \textbf{\begin{tabular}[c]{@{}c@{}}81.62\\ 0.7475\end{tabular}}}\\
\Xhline{3\arrayrulewidth}

\end{tabular}%
}
\caption{Experimental results of PiFi on different sizes of LM within same model family. For this experiment, we extracted the last layer of 0.5B, 1.5B, 7B version of Qwen2 as $L_{\text{LM}}$. We report accuracy and F1-score in the upper row and lower row of each cell.}
\label{tab:tab6-size}
\end{table}

\subsection{Comparison of PiFi between Various LLMs}
\label{sec:experiment-llm}

LLMs vary significantly in architecture and pre-training data, which can influence their performance as backbones for the PiFi framework. In this section, we evaluate the downstream task performance of PiFi when using different LLMs as sources for extracting $L_{\textit{LLM}}$. Specifically, we incorporate the following LLMs: Llama-3.1-8B \cite{dubey2024llama}, Mistral-7B-v0.1 and v0.3 \cite{jiang2023mistral}, Qwen2-7B \cite{yang2024qwen2}, Gemma-2-9B \cite{team2024gemma}, and Falcon-7B \cite{almazrouei2023falcon}. These LLMs serve as backbones for generating $L_{\textit{LLM}}$, which is subsequently integrated into a BERT model for the fine-tuning on text classification tasks.

Table~\ref{tab:tab5-llm} presents the results of these experiments. Our findings reveal several key insights. First, across all LLM variants, PiFi consistently outperforms the BERT model with vanilla fine-tuning, demonstrating the effectiveness of utilizing $L_{\textit{LLM}}$. Second, Llama-3.1-8B and Mistral-7B-v0.3 yield the highest performance across multiple datasets, indicating their strong suitability as backbones for PiFi. Additionally, the comparison between PiFi models using Mistral-7B-v0.1 and v0.3 shows that even incremental improvements within the same model family lead to enhanced downstream performance. This result suggests that PiFi is highly responsive to the advancements in LLM capabilities, and we expect that future advancements in LLM development will further augment the effectiveness of PiFi.

\subsection{Impact on PiFi Depending on LM Size}
\label{sec:experiment-size}

Currently, LMs are available in various sizes to cater to different computational and performance needs. In this section, we evaluate how the size of the LM used to extract LM layer affects the overall performance of PiFi. For this experiment, we utilize three versions of the Qwen2 model \cite{yang2024qwen2}: Qwen2-0.5B, Qwen2-1.5B, and Qwen2-7B.

Table~\ref{tab:tab6-size} demonstrates the results. We observe that integrating LM layer from any LM size into a SLM results in improved performance compared to traditional fine-tuning methods. Notably, the usage of 7B model yields the highest performance gains, surpassing both the 0.5B and 1.5B models. This suggests that larger models, with their enhanced capacity for capturing and storing extensive knowledge, provide more informative and effective representations for PiFi, leading to better downstream task performance.

\section{Conclusion}
\label{sec:conclusion}
In this paper, we introduced PiFi, plug-in and fine-tuning, a novel framework designed to leverage the intrinsic knowledge of LLMs while maintaining the efficiency and lightweight nature of SLMs. PiFi incorporates a frozen LLM layer into a SLM and fine-tunes the resulting model, enabling the effective utilization of LLM knowledge without significantly increasing model complexity.

We validated the applicability of PiFi across a diverse range of NLU and NLG tasks, demonstrating its compatibility with various SLMs and LLMs. Notably, the results from our domain shift experiment in Section~\ref{sec:experiment-domain} showed that PiFi can significantly improve the performance of SLMs on unseen domains. Similarly, the analysis in Section~\ref{sec:experiment-linguistic} confirmed that PiFi effectively leverages the linguistic capabilities of LLMs by incorporating even a single LLM layer.

In future work, we aim to extend the usability of PiFi to more tasks and languages. We will also explore advanced strategies to further optimize PiFi’s effectiveness, such as automatically selecting the number and position of LLM layers to be integrated, allowing for more flexible and task-specific knowledge transfer.

\section*{Limitations}
In this study, we proposed PiFi, an efficient framework for integrating the knowledge of LLMs into SLMs. Despite its effectiveness, there are several limitations in our current approach that warrant further exploration.

One limitation is that we selected $L_{\textit{LLM}}$ as the last layer of each LLM based on a heuristic approach. While our analysis in Appendix~\ref{app:ablation-layer-position} indicates that using the last layer generally yields the best performance, it is possible that different layers may be optimal depending on the specific downstream task. Future research could explore methods for automatically selecting the most suitable LLM layer as $L_{\textit{LLM}}$, akin to techniques used in neural architecture search \cite{elsken2019neural, white2023neural}.

Another limitation is that our experiments primarily focused on relatively straightforward tasks, as this manuscript’s primary goal is to propose PiFi and clearly demonstrate its effectiveness. We did not extend our evaluation to more complex benchmarks, such as the massive multitask language understanding dataset \cite{hendrycks2020measuring}. Testing PiFi on such diverse and challenging tasks in future work could provide deeper insights into its generalizability and highlight areas for further improvement.

\section*{Ethics Statement}
We acknowledge the potential for inherent biases in LLMs, and the integration of an LLM layer into a SLM may introduce such biases \cite{liu2022quantifying}. While our experiments did not reveal any explicit evidence of bias, future work will carefully consider the possibility of bias transfer from LLMs to SLMs when employing PiFi, as well as its broader implications.

\section*{Acknowledgements}
This work was supported by the Institute of Information \& Communications Technology Planning \& Evaluation (IITP) grant funded by the Korea government (MSIT) [RS-2021-II211341, Artificial Intelligent Graduate School Program (Chung-Ang University)] and by the National Research Foundation of Korea (NRF) grant funded by the Korea government (MSIT) (RS-2025-00556246).


\bibliography{custom}

\begin{thebibliography}{83}
\expandafter\ifx\csname natexlab\endcsname\relax\def\natexlab#1{#1}\fi

\bibitem[{Achiam et~al.(2023)Achiam, Adler, Agarwal, Ahmad, Akkaya, Aleman, Almeida, Altenschmidt, Altman, Anadkat et~al.}]{achiam2023gpt}
Josh Achiam, Steven Adler, Sandhini Agarwal, Lama Ahmad, Ilge Akkaya, Florencia~Leoni Aleman, Diogo Almeida, Janko Altenschmidt, Sam Altman, Shyamal Anadkat, et~al. 2023.
\newblock \href {https://arxiv.org/abs/2303.08774} {Gpt-4 technical report}.
\newblock \emph{arXiv preprint arXiv:2303.08774}.

\bibitem[{Allal et~al.(2025)Allal, Lozhkov, Bakouch, Bl{\'a}zquez, Penedo, Tunstall, Marafioti, Kydl{\'\i}{\v{c}}ek, Lajar{\'\i}n, Srivastav et~al.}]{allal2025smollm2}
Loubna~Ben Allal, Anton Lozhkov, Elie Bakouch, Gabriel~Mart{\'\i}n Bl{\'a}zquez, Guilherme Penedo, Lewis Tunstall, Andr{\'e}s Marafioti, Hynek Kydl{\'\i}{\v{c}}ek, Agust{\'\i}n~Piqueres Lajar{\'\i}n, Vaibhav Srivastav, et~al. 2025.
\newblock \href {https://arxiv.org/abs/2502.02737} {Smollm2: When smol goes big--data-centric training of a small language model}.
\newblock \emph{arXiv preprint arXiv:2502.02737}.

\bibitem[{Almazrouei et~al.(2023)Almazrouei, Alobeidli, Alshamsi, Cappelli, Cojocaru, Debbah, Goffinet, Hesslow, Launay, Malartic et~al.}]{almazrouei2023falcon}
Ebtesam Almazrouei, Hamza Alobeidli, Abdulaziz Alshamsi, Alessandro Cappelli, Ruxandra Cojocaru, M{\'e}rouane Debbah, {\'E}tienne Goffinet, Daniel Hesslow, Julien Launay, Quentin Malartic, et~al. 2023.
\newblock \href {https://arxiv.org/abs/2311.16867} {The falcon series of open language models}.
\newblock \emph{arXiv preprint arXiv:2311.16867}.

\bibitem[{Artzy and Schwartz(2024)}]{artzy2024attend}
Amit Artzy and Roy Schwartz. 2024.
\newblock \href {https://aclanthology.org/2024.blackboxnlp-1.10/} {Attend first, consolidate later: On the importance of attention in different llm layers}.
\newblock In \emph{Proceedings of EMNLP 2024 BlackboxNLP Workshop}, pages 177--184.

\bibitem[{Banerjee and Lavie(2005)}]{banerjee2005meteor}
Satanjeev Banerjee and Alon Lavie. 2005.
\newblock \href {https://aclanthology.org/W05-0909} {{METEOR}: An automatic metric for {MT} evaluation with improved correlation with human judgments}.
\newblock In \emph{Proceedings of ACL 2005 Workshop on Intrinsic and Extrinsic Evaluation Measures for Machine Translation and/or Summarization}, pages 65--72.

\bibitem[{Barbieri et~al.(2020)Barbieri, Camacho-Collados, Espinosa~Anke, and Neves}]{barbieri2020tweeteval}
Francesco Barbieri, Jose Camacho-Collados, Luis Espinosa~Anke, and Leonardo Neves. 2020.
\newblock \href {https://aclanthology.org/2020.findings-emnlp.148} {{T}weet{E}val: Unified benchmark and comparative evaluation for tweet classification}.
\newblock In \emph{Findings of EMNLP}, pages 1644--1650.

\bibitem[{Bowman et~al.(2015)Bowman, Angeli, Potts, and Manning}]{bowman2015large}
Samuel~R. Bowman, Gabor Angeli, Christopher Potts, and Christopher~D. Manning. 2015.
\newblock \href {https://aclanthology.org/D15-1075} {A large annotated corpus for learning natural language inference}.
\newblock In \emph{Proceedings of EMNLP}, pages 632--642.

\bibitem[{Brown et~al.(2020)Brown, Mann, Ryder, Subbiah, Kaplan, Dhariwal, Neelakantan, Shyam, Sastry, Askell et~al.}]{brown2020language}
Tom Brown, Benjamin Mann, Nick Ryder, Melanie Subbiah, Jared~D Kaplan, Prafulla Dhariwal, Arvind Neelakantan, Pranav Shyam, Girish Sastry, Amanda Askell, et~al. 2020.
\newblock \href {https://papers.nips.cc/paper/2020/hash/1457c0d6bfcb4967418bfb8ac142f64a-Abstract.html} {Language models are few-shot learners}.
\newblock In \emph{Proceedings of NeurIPS}, pages 1877--1901.

\bibitem[{Choi et~al.(2024)Choi, Kim, Yu, Yun, and Kim}]{choi2024unigen}
Juhwan Choi, Yeonghwa Kim, Seunguk Yu, Jungmin Yun, and Youngbin Kim. 2024.
\newblock \href {https://aclanthology.org/2024.emnlp-main.1/} {Unigen: Universal domain generalization for sentiment classification via zero-shot dataset generation}.
\newblock In \emph{Proceedings of EMNLP}, pages 1--14.

\bibitem[{Clark et~al.(2020)Clark, Luong, Le, and Manning}]{clark2020electra}
Kevin Clark, Minh-Thang Luong, Quoc~V. Le, and Christopher~D. Manning. 2020.
\newblock \href {https://openreview.net/forum?id=r1xMH1BtvB} {Electra: Pre-training text encoders as discriminators rather than generators}.
\newblock In \emph{Proceedings of ICLR}.

\bibitem[{Dettmers et~al.(2023)Dettmers, Pagnoni, Holtzman, and Zettlemoyer}]{dettmers2023qlora}
Tim Dettmers, Artidoro Pagnoni, Ari Holtzman, and Luke Zettlemoyer. 2023.
\newblock \href {https://proceedings.neurips.cc/paper_files/paper/2023/hash/1feb87871436031bdc0f2beaa62a049b-Abstract-Conference.html} {Qlora: Efficient finetuning of quantized llms}.
\newblock In \emph{Proceedings of NeurIPS}, pages 10088--10115.

\bibitem[{Devlin et~al.(2019)Devlin, Chang, Lee, and Toutanova}]{devlin2019bert}
Jacob Devlin, Ming-Wei Chang, Kenton Lee, and Kristina Toutanova. 2019.
\newblock \href {https://aclanthology.org/N19-1423} {{BERT}: Pre-training of deep bidirectional transformers for language understanding}.
\newblock In \emph{Proceedings of NAACL}, pages 4171--4186.

\bibitem[{Ding et~al.(2008)Ding, Liu, and Yu}]{ding2008holistic}
Xiaowen Ding, Bing Liu, and Philip~S Yu. 2008.
\newblock \href {https://dl.acm.org/doi/10.1145/1341531.1341561} {A holistic lexicon-based approach to opinion mining}.
\newblock In \emph{Proceedings of WSDM}, pages 231--240.

\bibitem[{DiscoResearch(2024)}]{llama3german}
DiscoResearch. 2024.
\newblock \href {https://huggingface.co/DiscoResearch/Llama3-German-8B} {Llama-3-german-8b}.
\newblock Hugging Face Repository.

\bibitem[{Dosovitskiy et~al.(2021)Dosovitskiy, Beyer, Kolesnikov, Weissenborn, Zhai, Unterthiner, Dehghani, Minderer, Heigold, Gelly, Uszkoreit, and Houlsby}]{dosovitskiy2021an}
Alexey Dosovitskiy, Lucas Beyer, Alexander Kolesnikov, Dirk Weissenborn, Xiaohua Zhai, Thomas Unterthiner, Mostafa Dehghani, Matthias Minderer, Georg Heigold, Sylvain Gelly, Jakob Uszkoreit, and Neil Houlsby. 2021.
\newblock \href {https://openreview.net/forum?id=YicbFdNTTy} {An image is worth 16x16 words: Transformers for image recognition at scale}.
\newblock In \emph{Proceedings of ICLR}.

\bibitem[{Dubey et~al.(2024)Dubey, Jauhri, Pandey, Kadian, Al-Dahle, Letman, Mathur, Schelten, Yang, Fan et~al.}]{dubey2024llama}
Abhimanyu Dubey, Abhinav Jauhri, Abhinav Pandey, Abhishek Kadian, Ahmad Al-Dahle, Aiesha Letman, Akhil Mathur, Alan Schelten, Amy Yang, Angela Fan, et~al. 2024.
\newblock \href {https://arxiv.org/abs/2407.21783} {The llama 3 herd of models}.
\newblock \emph{arXiv preprint arXiv:2407.21783}.

\bibitem[{Elliott et~al.(2016)Elliott, Frank, Sima'an, and Specia}]{elliott2016multi30k}
Desmond Elliott, Stella Frank, Khalil Sima'an, and Lucia Specia. 2016.
\newblock \href {https://aclanthology.org/W16-3210} {Multi30k: Multilingual english-german image descriptions}.
\newblock In \emph{Proceedings of ACL 2016 Workshop on Vision and Language}, pages 70--74.

\bibitem[{Elsken et~al.(2019)Elsken, Metzen, and Hutter}]{elsken2019neural}
Thomas Elsken, Jan~Hendrik Metzen, and Frank Hutter. 2019.
\newblock \href {https://www.jmlr.org/papers/v20/18-598.html} {Neural architecture search: A survey}.
\newblock \emph{Journal of Machine Learning Research}, 20(55):1--21.

\bibitem[{Fan et~al.(2023)Fan, Krishnan, Isola, Katabi, and Tian}]{fan2023improving}
Lijie Fan, Dilip Krishnan, Phillip Isola, Dina Katabi, and Yonglong Tian. 2023.
\newblock \href {https://proceedings.neurips.cc/paper_files/paper/2023/hash/6fa4d985e7c434002fb6289ab9b2d654-Abstract-Conference.html} {Improving clip training with language rewrites}.
\newblock In \emph{Proceedings of NeurIPS}, pages 35544--35575.

\bibitem[{Gao et~al.(2023{\natexlab{a}})Gao, Pi, Yong, Xu, Ye, Wu, Zhang, Liang, Li, and Kong}]{gao2023self}
Jiahui Gao, Renjie Pi, Lin Yong, Hang Xu, Jiacheng Ye, Zhiyong Wu, Weizhong Zhang, Xiaodan Liang, Zhenguo Li, and Lingpeng Kong. 2023{\natexlab{a}}.
\newblock \href {https://openreview.net/forum?id=h5OpjGd_lo6} {Self-guided noise-free data generation for efficient zero-shot learning}.
\newblock In \emph{Proceedings of ICLR}.

\bibitem[{Gao et~al.(2023{\natexlab{b}})Gao, Zhou, Liu, Zhao, and Wen}]{gao2023small}
Ze-Feng Gao, Kun Zhou, Peiyu Liu, Wayne~Xin Zhao, and Ji-Rong Wen. 2023{\natexlab{b}}.
\newblock \href {https://aclanthology.org/2023.acl-long.212} {Small pre-trained language models can be fine-tuned as large models via over-parameterization}.
\newblock In \emph{Proceedings of ACL}, pages 3819--3834.

\bibitem[{Gu et~al.(2024)Gu, Dong, Wei, and Huang}]{gu2024minillm}
Yuxian Gu, Li~Dong, Furu Wei, and Minlie Huang. 2024.
\newblock \href {https://openreview.net/forum?id=5h0qf7IBZZ} {Minillm: Knowledge distillation of large language models}.
\newblock In \emph{Proceedings of ICLR}.

\bibitem[{Guhr et~al.(2020)Guhr, Schumann, Bahrmann, and B{\"o}hme}]{guhr2020training}
Oliver Guhr, Anne-Kathrin Schumann, Frank Bahrmann, and Hans~Joachim B{\"o}hme. 2020.
\newblock \href {https://aclanthology.org/2020.lrec-1.202} {Training a broad-coverage {G}erman sentiment classification model for dialog systems}.
\newblock In \emph{Proceedings of LREC}, pages 1627--1632.

\bibitem[{Gururangan et~al.(2020)Gururangan, Marasovi{\'c}, Swayamdipta, Lo, Beltagy, Downey, and Smith}]{gururangan2020dont}
Suchin Gururangan, Ana Marasovi{\'c}, Swabha Swayamdipta, Kyle Lo, Iz~Beltagy, Doug Downey, and Noah~A. Smith. 2020.
\newblock \href {https://aclanthology.org/2020.acl-main.740} {Don’t stop pretraining: Adapt language models to domains and tasks}.
\newblock In \emph{Proceedings of ACL}, pages 8342--8360.

\bibitem[{He et~al.(2023)He, Gao, and Chen}]{he2023debertav}
Pengcheng He, Jianfeng Gao, and Weizhu Chen. 2023.
\newblock \href {https://openreview.net/forum?id=sE7-XhLxHA} {Debertav3: Improving deberta using electra-style pre-training with gradient-disentangled embedding sharing}.
\newblock In \emph{Proceedings of ICLR}.

\bibitem[{He et~al.(2021)He, Liu, Gao, and Chen}]{he2021deberta}
Pengcheng He, Xiaodong Liu, Jianfeng Gao, and Weizhu Chen. 2021.
\newblock \href {https://openreview.net/forum?id=XPZIaotutsD} {Deberta: Decoding-enhanced bert with disentangled attention}.
\newblock In \emph{Proceedings of ICLR}.

\bibitem[{Hendrycks et~al.(2021)Hendrycks, Burns, Basart, Zou, Mazeika, Song, and Steinhardt}]{hendrycks2020measuring}
Dan Hendrycks, Collin Burns, Steven Basart, Andy Zou, Mantas Mazeika, Dawn Song, and Jacob Steinhardt. 2021.
\newblock \href {https://openreview.net/forum?id=d7KBjmI3GmQ} {Measuring massive multitask language understanding}.
\newblock In \emph{Proceedings of ICLR}.

\bibitem[{Hu et~al.(2022)Hu, yelong shen, Wallis, Allen-Zhu, Li, Wang, Wang, and Chen}]{hu2022lora}
Edward~J Hu, yelong shen, Phillip Wallis, Zeyuan Allen-Zhu, Yuanzhi Li, Shean Wang, Lu~Wang, and Weizhu Chen. 2022.
\newblock \href {https://openreview.net/forum?id=nZeVKeeFYf9} {Lo{RA}: Low-rank adaptation of large language models}.
\newblock In \emph{Proceedings of ICLR}.

\bibitem[{Jiang et~al.(2023)Jiang, Sablayrolles, Mensch, Bamford, Chaplot, Casas, Bressand, Lengyel, Lample, Saulnier et~al.}]{jiang2023mistral}
Albert~Q. Jiang, Alexandre Sablayrolles, Arthur Mensch, Chris Bamford, Devendra~Singh Chaplot, Diego de~las Casas, Florian Bressand, Gianna Lengyel, Guillaume Lample, Lucile Saulnier, et~al. 2023.
\newblock \href {https://arxiv.org/abs/2310.06825} {Mistral 7b}.
\newblock \emph{arXiv preprint arXiv:2310.06825}.

\bibitem[{Jiao et~al.(2020)Jiao, Yin, Shang, Jiang, Chen, Li, Wang, and Liu}]{jiao2020tinybert}
Xiaoqi Jiao, Yichun Yin, Lifeng Shang, Xin Jiang, Xiao Chen, Linlin Li, Fang Wang, and Qun Liu. 2020.
\newblock \href {https://aclanthology.org/2020.findings-emnlp.372} {{T}iny{BERT}: Distilling {BERT} for natural language understanding}.
\newblock In \emph{Findings of EMNLP}, pages 4163--4174.

\bibitem[{Ju et~al.(2024)Ju, Sun, Du, Yuan, Ren, and Liu}]{ju2024large}
Tianjie Ju, Weiwei Sun, Wei Du, Xinwei Yuan, Zhaochun Ren, and Gongshen Liu. 2024.
\newblock \href {https://aclanthology.org/2024.lrec-main.722} {How large language models encode context knowledge? a layer-wise probing study}.
\newblock In \emph{Proceedings of LREC-COLING}, pages 8235--8246.

\bibitem[{Kingma and Ba(2015)}]{kingma2014adam}
Diederik~P. Kingma and Jimmy Ba. 2015.
\newblock \href {https://arxiv.org/abs/1412.6980} {Adam: A method for stochastic optimization}.
\newblock In \emph{Proceedings of ICLR}.

\bibitem[{Lee(2024)}]{llama3openko}
Junbum Lee. 2024.
\newblock \href {https://huggingface.co/beomi/Llama-3-Open-Ko-8B} {Llama-3-open-ko}.
\newblock Hugging Face Repository.

\bibitem[{Lepagnol et~al.(2024)Lepagnol, Gerald, Ghannay, Servan, and Rosset}]{lepagnol2024small}
Pierre Lepagnol, Thomas Gerald, Sahar Ghannay, Christophe Servan, and Sophie Rosset. 2024.
\newblock \href {https://aclanthology.org/2024.lrec-main.1299} {Small language models are good too: An empirical study of zero-shot classification}.
\newblock In \emph{Proceedings of LREC-COLING}, pages 14923--14936.

\bibitem[{Lewis et~al.(2020)Lewis, Liu, Goyal, Ghazvininejad, Mohamed, Levy, Stoyanov, and Zettlemoyer}]{lewis2020bart}
Mike Lewis, Yinhan Liu, Naman Goyal, Marjan Ghazvininejad, Abdelrahman Mohamed, Omer Levy, Veselin Stoyanov, and Luke Zettlemoyer. 2020.
\newblock \href {https://aclanthology.org/2020.acl-main.703} {{BART}: Denoising sequence-to-sequence pre-training for natural language generation, translation, and comprehension}.
\newblock In \emph{Proceedings of ACL}, pages 7871--7880.

\bibitem[{Lin(2004)}]{lin2004rouge}
Chin-Yew Lin. 2004.
\newblock \href {https://aclanthology.org/W04-1013/} {Rouge: A package for automatic evaluation of summaries}.
\newblock In \emph{Proceedings of ACL 2004 Workshop Text Summarization Branches Out}, pages 74--81.

\bibitem[{Lin et~al.(2023)Lin, Qu, Chen, Chen, Chen, and Huang}]{lin2023pushing}
Zheng Lin, Guanqiao Qu, Qiyuan Chen, Xianhao Chen, Zhe Chen, and Kaibin Huang. 2023.
\newblock \href {https://openreview.net/forum?id=VF8OGR9skB} {Pushing large language models to the 6g edge: Vision, challenges, and opportunities}.
\newblock \emph{CoRR}.

\bibitem[{Liu et~al.(2022)Liu, Jia, Wei, Xu, and Vosoughi}]{liu2022quantifying}
Ruibo Liu, Chenyan Jia, Jason Wei, Guangxuan Xu, and Soroush Vosoughi. 2022.
\newblock \href {https://www.sciencedirect.com/science/article/pii/S0004370221002058?dgcid=coauthor} {Quantifying and alleviating political bias in language models}.
\newblock \emph{Artificial Intelligence}, 304:103654.

\bibitem[{Liu et~al.(2019)Liu, Ott, Goyal, Du, Joshi, Chen, Levy, Lewis, Zettlemoyer, and Stoyanov}]{liu2019roberta}
Yinhan Liu, Myle Ott, Naman Goyal, Jingfei Du, Mandar Joshi, Danqi Chen, Omer Levy, Mike Lewis, Luke Zettlemoyer, and Veselin Stoyanov. 2019.
\newblock \href {https://arxiv.org/abs/1907.11692} {Roberta: A robustly optimized {BERT} pretraining approach}.
\newblock \emph{arXiv preprint arXiv:1907.11692}.

\bibitem[{Liu et~al.(2024)Liu, Kong, Liu, and Sun}]{liu2024fantastic}
Zhu Liu, Cunliang Kong, Ying Liu, and Maosong Sun. 2024.
\newblock \href {https://aclanthology.org/2024.findings-acl.866} {Fantastic semantics and where to find them: Investigating which layers of generative {LLM}s reflect lexical semantics}.
\newblock In \emph{Findings of ACL}, pages 14551--14558.

\bibitem[{Lu et~al.(2024)Lu, Li, Cai, Yi, Liu, Zhang, Lane, and Xu}]{lu2024small}
Zhenyan Lu, Xiang Li, Dongqi Cai, Rongjie Yi, Fangming Liu, Xiwen Zhang, Nicholas~D Lane, and Mengwei Xu. 2024.
\newblock \href {https://arxiv.org/abs/2409.15790} {Small language models: Survey, measurements, and insights}.
\newblock \emph{arXiv preprint arXiv:2409.15790}.

\bibitem[{Luo et~al.(2023)Luo, Yang, Meng, Li, Zhou, and Zhang}]{luo2023empirical}
Yun Luo, Zhen Yang, Fandong Meng, Yafu Li, Jie Zhou, and Yue Zhang. 2023.
\newblock \href {https://arxiv.org/abs/2308.08747} {An empirical study of catastrophic forgetting in large language models during continual fine-tuning}.
\newblock \emph{arXiv preprint arXiv:2308.08747}.

\bibitem[{Maas et~al.(2011)Maas, Daly, Pham, Huang, Ng, and Potts}]{maas2011learning}
Andrew~L. Maas, Raymond~E. Daly, Peter~T. Pham, Dan Huang, Andrew~Y. Ng, and Christopher Potts. 2011.
\newblock \href {https://aclanthology.org/P11-1015} {Learning word vectors for sentiment analysis}.
\newblock In \emph{Proceedings of ACL}, pages 142--150.

\bibitem[{Meta(2019)}]{Meta2019fvcore}
Meta. 2019.
\newblock \href {https://github.com/facebookresearch/fvcore} {fvcore}.
\newblock GitHub Repository.

\bibitem[{Meta(2024)}]{llama31english}
Meta. 2024.
\newblock \href {https://huggingface.co/meta-llama/Llama-3.1-8B} {Llama-3.1-8b}.
\newblock Hugging Face Repository.

\bibitem[{Minaee et~al.(2024)Minaee, Mikolov, Nikzad, Chenaghlu, Socher, Amatriain, and Gao}]{minaee2024large}
Shervin Minaee, Tomas Mikolov, Narjes Nikzad, Meysam Chenaghlu, Richard Socher, Xavier Amatriain, and Jianfeng Gao. 2024.
\newblock \href {https://arxiv.org/abs/2402.06196} {Large language models: A survey}.
\newblock \emph{arXiv preprint arXiv:2402.06196}.

\bibitem[{Nallapati et~al.(2016)Nallapati, Zhou, dos Santos, G{\.{u}}l{\c{c}}ehre, and Xiang}]{nallapati2016abstractive}
Ramesh Nallapati, Bowen Zhou, Cicero dos Santos, {\c{C}}a{\u{g}}lar G{\.{u}}l{\c{c}}ehre, and Bing Xiang. 2016.
\newblock \href {https://aclanthology.org/K16-1028} {Abstractive text summarization using sequence-to-sequence {RNN}s and beyond}.
\newblock In \emph{Proceedings of SIGNLL}, pages 280--290.

\bibitem[{Nityasya et~al.(2020)Nityasya, Wibowo, Prasojo, and Aji}]{nityasya2020costs}
Made~Nindyatama Nityasya, Haryo~Akbarianto Wibowo, Radityo~Eko Prasojo, and Alham~Fikri Aji. 2020.
\newblock \href {https://arxiv.org/abs/2012.08958} {Costs to consider in adopting nlp for your business}.
\newblock \emph{arXiv preprint arXiv:2012.08958}.

\bibitem[{Ouyang et~al.(2022)Ouyang, Wu, Jiang, Almeida, Wainwright, Mishkin, Zhang, Agarwal, Slama, Ray et~al.}]{ouyang2022training}
Long Ouyang, Jeffrey Wu, Xu~Jiang, Diogo Almeida, Carroll Wainwright, Pamela Mishkin, Chong Zhang, Sandhini Agarwal, Katarina Slama, Alex Ray, et~al. 2022.
\newblock \href {https://proceedings.neurips.cc/paper_files/paper/2022/hash/b1efde53be364a73914f58805a001731-Abstract-Conference.html} {Training language models to follow instructions with human feedback}.
\newblock In \emph{Proceedings of NeurIPS}, pages 27730--27744.

\bibitem[{Pang et~al.(2024)Pang, Xie, Man, and Wang}]{pang2024frozen}
Ziqi Pang, Ziyang Xie, Yunze Man, and Yu-Xiong Wang. 2024.
\newblock \href {https://openreview.net/forum?id=t0FI3Q66K5} {Frozen transformers in language models are effective visual encoder layers}.
\newblock In \emph{Proceedings of ICLR}.

\bibitem[{Papineni et~al.(2002)Papineni, Roukos, Ward, and Zhu}]{papineni2002bleu}
Kishore Papineni, Salim Roukos, Todd Ward, and Wei-Jing Zhu. 2002.
\newblock \href {https://aclanthology.org/P02-1040} {Bleu: A method for automatic evaluation of machine translation}.
\newblock In \emph{Proceedings of ACL}, pages 311--318.

\bibitem[{Park(2016)}]{park2016nsmc}
Lucy Park. 2016.
\newblock \href {https://github.com/e9t/nsmc/} {Naver sentiment movie corpus}.
\newblock GitHub Repository.

\bibitem[{Paszke et~al.(2019{\natexlab{a}})Paszke, Gross, Massa, Lerer, Bradbury, Chanan, Killeen, Lin, Gimelshein, Antiga, Desmaison, Kopf, Yang, DeVito, Raison, Tejani, Chilamkurthy, Steiner, Fang, Bai, and Chintala}]{paszke2019pytorchimperativestylehighperformance}
Adam Paszke, Sam Gross, Francisco Massa, Adam Lerer, James Bradbury, Gregory Chanan, Trevor Killeen, Zeming Lin, Natalia Gimelshein, Luca Antiga, Alban Desmaison, Andreas Kopf, Edward Yang, Zachary DeVito, Martin Raison, Alykhan Tejani, Sasank Chilamkurthy, Benoit Steiner, Lu~Fang, Junjie Bai, and Soumith Chintala. 2019{\natexlab{a}}.
\newblock \href {https://papers.nips.cc/paper_files/paper/2019/hash/bdbca288fee7f92f2bfa9f7012727740-Abstract.html} {Pytorch: An imperative style, high-performance deep learning library}.
\newblock In \emph{Proceedings of NeurIPS}.

\bibitem[{Paszke et~al.(2019{\natexlab{b}})Paszke, Gross, Massa, Lerer, Bradbury, Chanan, Killeen, Lin, Gimelshein, Antiga et~al.}]{paszke2019pytorch}
Adam Paszke, Sam Gross, Francisco Massa, Adam Lerer, James Bradbury, Gregory Chanan, Trevor Killeen, Zeming Lin, Natalia Gimelshein, Luca Antiga, et~al. 2019{\natexlab{b}}.
\newblock \href {https://papers.nips.cc/paper_files/paper/2019/hash/bdbca288fee7f92f2bfa9f7012727740-Abstract.html} {Pytorch: An imperative style, high-performance deep learning library}.
\newblock In \emph{Proceedings of NeurIPS}.

\bibitem[{Pires et~al.(2019)Pires, Schlinger, and Garrette}]{pires2019multilingual}
Telmo Pires, Eva Schlinger, and Dan Garrette. 2019.
\newblock \href {https://aclanthology.org/P19-1493} {How multilingual is multilingual {BERT}?}
\newblock In \emph{Proceedings of ACL}, pages 4996--5001.

\bibitem[{Radford et~al.(2021)Radford, Kim, Hallacy, Ramesh, Goh, Agarwal, Sastry, Askell, Mishkin, Clark et~al.}]{radford2021learning}
Alec Radford, Jong~Wook Kim, Chris Hallacy, Aditya Ramesh, Gabriel Goh, Sandhini Agarwal, Girish Sastry, Amanda Askell, Pamela Mishkin, Jack Clark, et~al. 2021.
\newblock \href {https://proceedings.mlr.press/v139/radford21a} {Learning transferable visual models from natural language supervision}.
\newblock In \emph{Proceedings of ICML}, pages 8748--8763.

\bibitem[{Radford and Narasimhan(2018)}]{Radford2018ImprovingLU}
Alec Radford and Karthik Narasimhan. 2018.
\newblock \href {https://openai.com/index/language-unsupervised/} {Improving language understanding by generative pre-training}.
\newblock \emph{OpenAI Blog}.

\bibitem[{Raffel et~al.(2020)Raffel, Shazeer, Roberts, Lee, Narang, Matena, Zhou, Li, and Liu}]{raffel2020exploring}
Colin Raffel, Noam Shazeer, Adam Roberts, Katherine Lee, Sharan Narang, Michael Matena, Yanqi Zhou, Wei Li, and Peter~J Liu. 2020.
\newblock \href {https://jmlr.org/papers/v21/20-074.html} {Exploring the limits of transfer learning with a unified text-to-text transformer}.
\newblock \emph{Journal of Machine Learning Research}, 21(140):1--67.

\bibitem[{Rajpurkar et~al.(2016)Rajpurkar, Zhang, Lopyrev, and Liang}]{rajpurkar2016squad}
Pranav Rajpurkar, Jian Zhang, Konstantin Lopyrev, and Percy Liang. 2016.
\newblock \href {https://aclanthology.org/D16-1264} {{SQ}u{AD}: 100,000+ questions for machine comprehension of text}.
\newblock In \emph{Proceedings of EMNLP}, pages 2383--2392.

\bibitem[{Rosenthal et~al.(2017)Rosenthal, Farra, and Nakov}]{rosenthal2017semeval}
Sara Rosenthal, Noura Farra, and Preslav Nakov. 2017.
\newblock \href {https://aclanthology.org/S17-2088/} {Semeval-2017 task 4: Sentiment analysis in twitter}.
\newblock In \emph{Proceedings of ACL 2017 Workshop on Semantic Evaluation}, pages 502--518.

\bibitem[{Shashidhar et~al.(2023)Shashidhar, Chinta, Sahai, Wang, and Ji}]{shashidhar2023democratizing}
Sumuk Shashidhar, Abhinav Chinta, Vaibhav Sahai, Zhenhailong Wang, and Heng Ji. 2023.
\newblock \href {https://aclanthology.org/2023.findings-emnlp.608} {Democratizing llms: An exploration of cost-performance trade-offs in self-refined open-source models}.
\newblock In \emph{Findings of EMNLP}, pages 9070--9084.

\bibitem[{Socher et~al.(2013)Socher, Perelygin, Wu, Chuang, Manning, Ng, and Potts}]{socher2013recursive}
Richard Socher, Alex Perelygin, Jean Wu, Jason Chuang, Christopher~D. Manning, Andrew Ng, and Christopher Potts. 2013.
\newblock \href {https://aclanthology.org/D13-1170} {Recursive deep models for semantic compositionality over a sentiment treebank}.
\newblock In \emph{Proceedings of EMNLP}, pages 1631--1642.

\bibitem[{Team et~al.(2024)Team, Riviere, Pathak, Sessa, Hardin, Bhupatiraju, Hussenot, Mesnard, Shahriari, Ram{\'e} et~al.}]{team2024gemma}
Gemma Team, Morgane Riviere, Shreya Pathak, Pier~Giuseppe Sessa, Cassidy Hardin, Surya Bhupatiraju, L{\'e}onard Hussenot, Thomas Mesnard, Bobak Shahriari, Alexandre Ram{\'e}, et~al. 2024.
\newblock \href {https://arxiv.org/abs/2408.00118} {Gemma 2: Improving open language models at a practical size}.
\newblock \emph{arXiv preprint arXiv:2408.00118}.

\bibitem[{Timiryasov and Tastet(2023)}]{timiryasov-tastet-2023-baby}
Inar Timiryasov and Jean-Loup Tastet. 2023.
\newblock \href {https://aclanthology.org/2023.conll-babylm.24} {Baby llama: Knowledge distillation from an ensemble of teachers trained on a small dataset with no performance penalty}.
\newblock In \emph{Proceedings of CoNLL 2024 BabyLM Challenge}, pages 279--289.

\bibitem[{Vaswani et~al.(2017)Vaswani, Shazeer, Parmar, Uszkoreit, Jones, Gomez, Kaiser, and Polosukhin}]{vaswani2017attention}
Ashish Vaswani, Noam Shazeer, Niki Parmar, Jakob Uszkoreit, Llion Jones, Aidan~N Gomez, Łukasz Kaiser, and Illia Polosukhin. 2017.
\newblock \href {https://papers.nips.cc/paper_files/paper/2017/hash/3f5ee243547dee91fbd053c1c4a845aa-Abstract.html} {Attention is all you need}.
\newblock In \emph{Proceedings of NeurIPS}, volume~30.

\bibitem[{Wang et~al.(2023)Wang, Chen, Ge, Xia, Bao, Zheng, Zhang, Gui, and Huang}]{wang2023orthogonal}
Xiao Wang, Tianze Chen, Qiming Ge, Han Xia, Rong Bao, Rui Zheng, Qi~Zhang, Tao Gui, and Xuanjing Huang. 2023.
\newblock \href {https://aclanthology.org/2023.findings-emnlp.715} {Orthogonal subspace learning for language model continual learning}.
\newblock In \emph{Findings of EMNLP}, pages 10658--10671.

\bibitem[{Warstadt et~al.(2019)Warstadt, Singh, and Bowman}]{warstadt2019neural}
Alex Warstadt, Amanpreet Singh, and Samuel~R. Bowman. 2019.
\newblock \href {https://aclanthology.org/Q19-1040} {Neural network acceptability judgments}.
\newblock \emph{Transactions of the Association for Computational Linguistics}, 7:625--641.

\bibitem[{Wei et~al.(2022)Wei, Tay, Bommasani, Raffel, Zoph, Borgeaud, Yogatama, Bosma, Zhou, Metzler et~al.}]{wei2022emergent}
Jason Wei, Yi~Tay, Rishi Bommasani, Colin Raffel, Barret Zoph, Sebastian Borgeaud, Dani Yogatama, Maarten Bosma, Denny Zhou, Donald Metzler, et~al. 2022.
\newblock \href {https://openreview.net/forum?id=yzkSU5zdwD} {Emergent abilities of large language models}.
\newblock \emph{Transactions on Machine Learning Research}.

\bibitem[{West et~al.(2022)West, Bhagavatula, Hessel, Hwang, Jiang, Le~Bras, Lu, Welleck, and Choi}]{west2022symbolic}
Peter West, Chandra Bhagavatula, Jack Hessel, Jena Hwang, Liwei Jiang, Ronan Le~Bras, Ximing Lu, Sean Welleck, and Yejin Choi. 2022.
\newblock \href {https://aclanthology.org/2022.naacl-main.341} {Symbolic knowledge distillation: From general language models to commonsense models}.
\newblock In \emph{Proceedings of NAACL}, pages 4602--4625.

\bibitem[{White et~al.(2023)White, Safari, Sukthanker, Ru, Elsken, Zela, Dey, and Hutter}]{white2023neural}
Colin White, Mahmoud Safari, Rhea Sukthanker, Binxin Ru, Thomas Elsken, Arber Zela, Debadeepta Dey, and Frank Hutter. 2023.
\newblock \href {https://arxiv.org/abs/2301.08727} {Neural architecture search: Insights from 1000 papers}.
\newblock \emph{arXiv preprint arXiv:2301.08727}.

\bibitem[{Williams et~al.(2018)Williams, Nangia, and Bowman}]{williams2018broad}
Adina Williams, Nikita Nangia, and Samuel Bowman. 2018.
\newblock \href {https://aclanthology.org/N18-1101} {A broad-coverage challenge corpus for sentence understanding through inference}.
\newblock In \emph{Proceedings of NAACL}, pages 1112--1122.

\bibitem[{Wolf et~al.(2020)Wolf, Debut, Sanh, Chaumond, Delangue, Moi, Cistac, Rault, Louf, Funtowicz et~al.}]{wolf2020transformers}
Thomas Wolf, Lysandre Debut, Victor Sanh, Julien Chaumond, Clement Delangue, Anthony Moi, Pierric Cistac, Tim Rault, R{\'e}mi Louf, Morgan Funtowicz, et~al. 2020.
\newblock \href {https://aclanthology.org/2020.emnlp-demos.6/} {Transformers: State-of-the-art natural language processing}.
\newblock In \emph{Proceedings of EMNLP (Demo Track)}, pages 38--45.

\bibitem[{Yang et~al.(2024)Yang, Yang, Hui, Zheng, Yu, Zhou, Li, Li, Liu, Huang et~al.}]{yang2024qwen2}
An~Yang, Baosong Yang, Binyuan Hui, Bo~Zheng, Bowen Yu, Chang Zhou, Chengpeng Li, Chengyuan Li, Dayiheng Liu, Fei Huang, et~al. 2024.
\newblock \href {https://arxiv.org/abs/2407.10671} {Qwen2 technical report}.
\newblock \emph{arXiv preprint arXiv:2407.10671}.

\bibitem[{Ye et~al.(2022)Ye, Gao, Li, Xu, Feng, Wu, Yu, and Kong}]{ye2022zerogen}
Jiacheng Ye, Jiahui Gao, Qintong Li, Hang Xu, Jiangtao Feng, Zhiyong Wu, Tao Yu, and Lingpeng Kong. 2022.
\newblock \href {https://aclanthology.org/2022.emnlp-main.801/} {Zerogen: Efficient zero-shot learning via dataset generation}.
\newblock In \emph{Proceedings of EMNLP}, pages 11653--11669.

\bibitem[{Ye(2024)}]{ye2024cross}
Qinyuan Ye. 2024.
\newblock \href {https://aclanthology.org/2024.naacl-srw.27} {Cross-task generalization abilities of large language models}.
\newblock In \emph{Proceedings of NAACL 2024 Student Research Workshop}, pages 255--262.

\bibitem[{Yu et~al.(2023)Yu, Yang, Pelrine, Godbout, and Rabbany}]{yu2023open}
Hao Yu, Zachary Yang, Kellin Pelrine, Jean-François Godbout, and Reihaneh Rabbany. 2023.
\newblock \href {https://arxiv.org/abs/2308.10092} {Open, closed, or small language models for text classification?}
\newblock \emph{arXiv preprint arXiv:2308.10092}.

\bibitem[{Yuan et~al.(2021)Yuan, Neubig, and Liu}]{yuan2021bartscore}
Weizhe Yuan, Graham Neubig, and Pengfei Liu. 2021.
\newblock \href {https://proceedings.neurips.cc/paper/2021/hash/e4d2b6e6fdeca3e60e0f1a62fee3d9dd-Abstract.html} {Bartscore: Evaluating generated text as text generation}.
\newblock In \emph{Proceedings of NeurIPS}, pages 27263--27277.

\bibitem[{Zampieri et~al.(2019)Zampieri, Malmasi, Nakov, Rosenthal, Farra, and Kumar}]{zampieri2019semeval}
Marcos Zampieri, Shervin Malmasi, Preslav Nakov, Sara Rosenthal, Noura Farra, and Ritesh Kumar. 2019.
\newblock \href {https://aclanthology.org/S19-2010/} {Semeval-2019 task 6: Identifying and categorizing offensive language in social media (offenseval)}.
\newblock In \emph{Proceedings of NAACL 2019 Workshop on Semantic Evaluation}, pages 75--86.

\bibitem[{Zhang et~al.(2024{\natexlab{a}})Zhang, Han, Liu, Zhou, Lu, Qiao, Li, and Gao}]{zhang2024llamaadapter}
Renrui Zhang, Jiaming Han, Chris Liu, Aojun Zhou, Pan Lu, Yu~Qiao, Hongsheng Li, and Peng Gao. 2024{\natexlab{a}}.
\newblock \href {https://openreview.net/forum?id=d4UiXAHN2W} {{LL}a{MA}-adapter: Efficient fine-tuning of large language models with zero-initialized attention}.
\newblock In \emph{Proceedings of ICLR}.

\bibitem[{Zhang et~al.(2020)Zhang, Kishore, Wu, Weinberger, and Artzi}]{zhang2020bertscore}
Tianyi Zhang, Varsha Kishore, Felix Wu, Kilian~Q Weinberger, and Yoav Artzi. 2020.
\newblock \href {https://openreview.net/forum?id=SkeHuCVFDr} {Bertscore: Evaluating text generation with bert}.
\newblock In \emph{Proceedings of ICLR}.

\bibitem[{Zhang et~al.(2024{\natexlab{b}})Zhang, Dong, and Kawaguchi}]{zhang2024investigating}
Yang Zhang, Yanfei Dong, and Kenji Kawaguchi. 2024{\natexlab{b}}.
\newblock \href {https://aclanthology.org/2024.blackboxnlp-1.29/} {Investigating layer importance in large language models}.
\newblock In \emph{Proceedings of EMNLP 2024 BlackboxNLP Workshop}, pages 469--479.

\bibitem[{Zhao et~al.(2023)Zhao, Zhou, Li, Tang, Wang, Hou, Min, Zhang, Zhang, Dong et~al.}]{zhao2023survey}
Wayne~Xin Zhao, Kun Zhou, Junyi Li, Tianyi Tang, Xiaolei Wang, Yupeng Hou, Yingqian Min, Beichen Zhang, Junjie Zhang, Zican Dong, et~al. 2023.
\newblock \href {https://arxiv.org/abs/2303.18223} {A survey of large language models}.
\newblock \emph{arXiv preprint arXiv:2303.18223}.

\bibitem[{Zhong et~al.(2024)Zhong, An, Chen, Han, and He}]{zhong2024seeking}
Ming Zhong, Chenxin An, Weizhu Chen, Jiawei Han, and Pengcheng He. 2024.
\newblock \href {https://openreview.net/forum?id=mIEHIcHGOo} {Seeking neural nuggets: Knowledge transfer in large language models from a parametric perspective}.
\newblock In \emph{Proceedings of ICLR}.

\end{thebibliography}
\bibliographystyle{acl_natbib}

\appendix

\section{Implementation Detail}
\label{app:implementation-detail}

We implemented PiFi using the PyTorch \cite{paszke2019pytorch} and Transformers \cite{wolf2020transformers} libraries. For encoder-based LMs, we used the representation of the $\textit{CLS}$ token as $h_{\textit{enc}}$. Similarly, for encoder-decoder models, we used the hidden states of the last encoder layer for each token as $h_{\textit{enc}}$. Each model was trained for three epochs with a learning rate of 5e-5 and a batch size of 32 using the Adam optimizer \cite{kingma2014adam}. Table~\ref{tab:tab8-parameter} provides the number of parameters used for each PiFi configuration across various SLMs and LLMs. Please refer to attached source code for further details.\footnote{\href{https://github.com/khyun8072/PiFi}{https://github.com/khyun8072/PiFi}}

\section{Ablation Study}
\label{app:ablation}

In this section, we present the results of ablation studies to supplement the main experiments conducted in Section~\ref{sec:experiment}.

\subsection{Effect of Selection of Layers within LLM}
\label{app:ablation-layer-position}

For the experiments in Section~\ref{sec:experiment}, we consistently used the last layer of each LLM as $L_{\textit{LLM}}$. However, recent studies have shown that different layers of LLMs capture distinct types of information and serve different roles \cite{artzy2024attend, zhang2024investigating, liu2024fantastic}. To explore this, we analyzed the effectiveness of PiFi when extracting $L_{\textit{LLM}}$ from layers other than the last. Specifically, we experimented with layers ${1, 4, 8, 12, 16, 20, 24, 28, 32}$ of the LLMs used in Section~\ref{sec:experiment-llm}.

\begin{figure*}[t]
    \centering
    \includegraphics[width=0.9\textwidth]{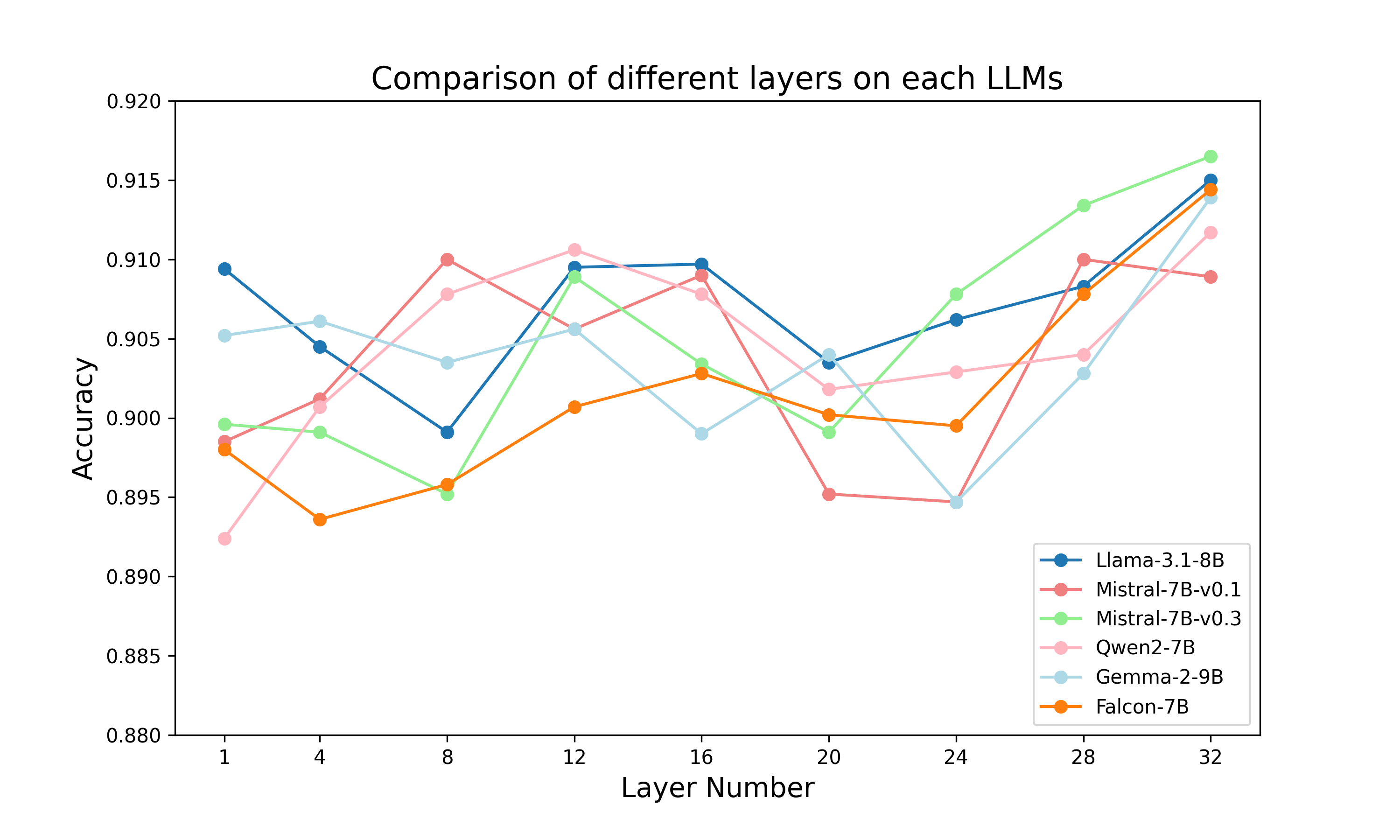}
    \caption{Comparison of the performance of PiFi models with different position of $L_{\textit{LLM}}$ on SST-2 dataset.}
\label{fig:layer-position}
\end{figure*}

Figure~\ref{fig:layer-position} presents the experimental results on the SST-2 dataset. The results clearly show that using the last layer yields the best performance compared to other layers, which aligns with previous findings suggesting that the upper layers of LLMs contain more contextual knowledge \cite{ju2024large}.

\begin{figure*}[t]
    \centering
    \includegraphics[width=0.9\textwidth]{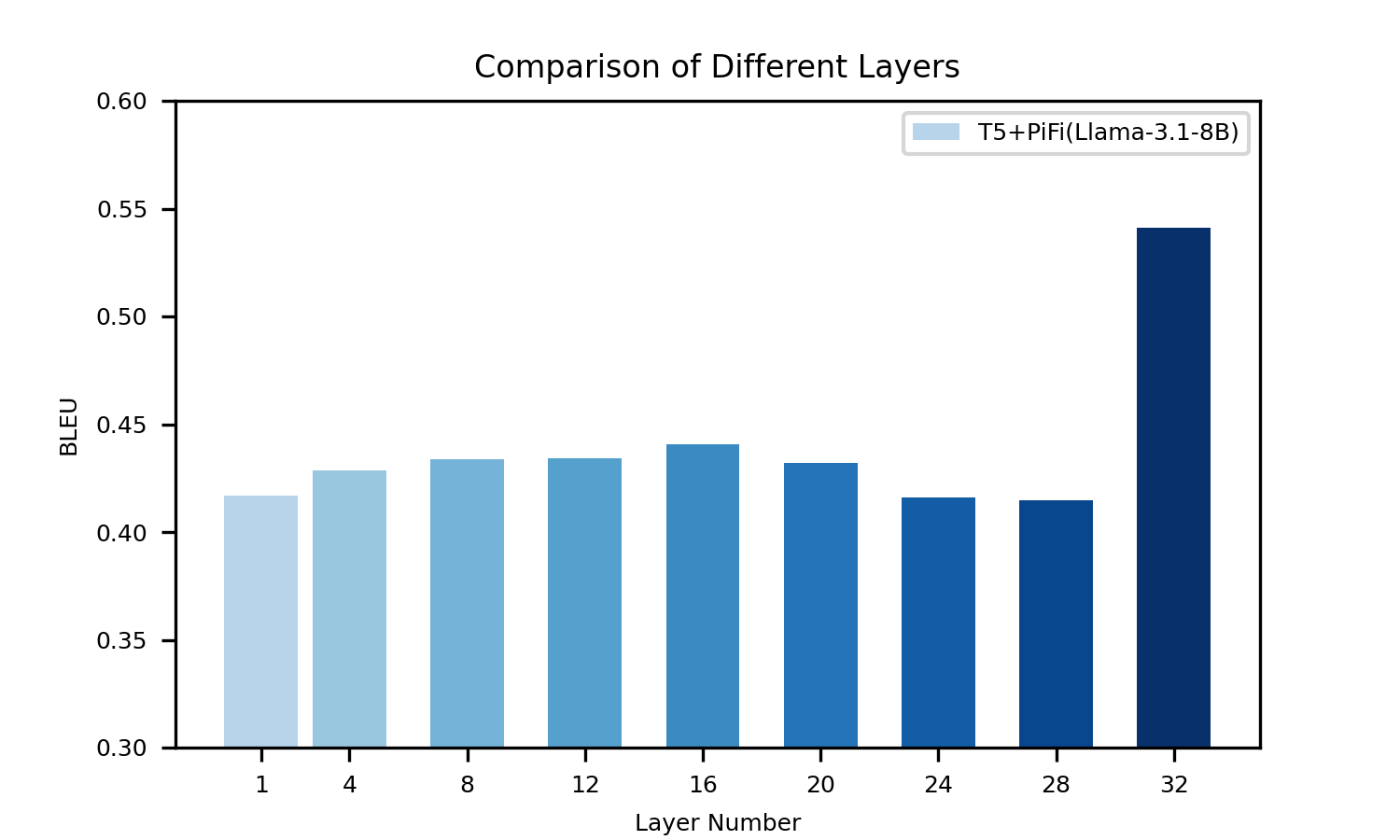}
    \caption{Comparison of the performance of PiFi models with different position of $L_{\textit{LLM}}$ on Multi30K dataset.}
\label{fig:layer-position-generation}
\end{figure*}

In addition to classification tasks, we also evaluated the effectiveness of PiFi in generation tasks. Specifically, we conducted a layer-wise performance analysis on the Multi30K translation task. As shown in Figure~\ref{fig:layer-position-generation}, the last layer consistently achieved the best performance in generation tasks. However, we acknowledge that for certain tasks, intermediate or other layers may be more suitable. This highlights the importance of task-specific optimization in layer selection. Therefore, in future research, we plan to explore optimization methods for selecting the most suitable layers for different tasks. Additionally, we aim to extend our work by leveraging automated techniques, such as Neural Architecture Search, to identify the optimal layers more efficiently.

\subsection{Performance of PiFi on Decoder-Based Models}
\label{app:decoder-base}
The PiFi framework has primarily been tested on encoder-based and encoder-decoder models. However, it is crucial to verify whether it exhibits the same effectiveness in decoder-based models (e.g., GPT-style architectures). To investigate its applicability and effectiveness further, we evaluated the performance of PiFi on a decoder-based model, SmolLM2-135M \cite{allal2025smollm2}. SmolLM2-135M is a relatively small decoder-based model, comparable in size to BERT-base. In this experiment, we added a classification head at the end of the model to enable it to perform various classification tasks. 

As shown in Table~\ref{tab:tab16-decoder model}, the decoder-based PiFi model outperforms the base SmolLM2-135M model across all tasks. This result indicates that PiFi is not limited to encoder-based and encoder-decoder models but can also be effectively applied to decoder-based architectures. These findings highlight the high versatility and scalability of PiFi, providing strong evidence for its applicability across a wide range of language model architectures.

% \usepackage{graphicx}
\begin{table}[t]
\centering
\resizebox{\columnwidth}{!}{%
\begin{tabular}{l|ccccc}
\Xhline{5\arrayrulewidth}
 &
  SST-2 &
  IMDB &
  \begin{tabular}[c]{@{}c@{}}Tweet\\ (Sentiment)\end{tabular} &
  \begin{tabular}[c]{@{}c@{}}Tweet\\ (Offensive)\end{tabular} &
  CoLA \\ \hline\hline
SmolLM2-135M &
  \begin{tabular}[c]{@{}c@{}}92.48\\ 0.9224\end{tabular} &
  \begin{tabular}[c]{@{}c@{}}87.70\\ 0.4817\end{tabular} &
  \begin{tabular}[c]{@{}c@{}}90.47\\ 0.8980\end{tabular} &
  \begin{tabular}[c]{@{}c@{}}84.42\\ 0.7673\end{tabular} &
  \begin{tabular}[c]{@{}c@{}}78.93\\ 0.7111\end{tabular} \\ \hline
\textbf{\begin{tabular}[c]{@{}l@{}}+PiFi\\ (Llama-3.1-8B)\end{tabular}} &
  \textbf{\begin{tabular}[c]{@{}c@{}}93.42\\ 0.9319\end{tabular}} &
  \textbf{\begin{tabular}[c]{@{}c@{}}88.69\\ 0.4955\end{tabular}} &
  \textbf{\begin{tabular}[c]{@{}c@{}}92.20\\ 0.9160\end{tabular}} &
  \textbf{\begin{tabular}[c]{@{}c@{}}86.04\\ 0.7967\end{tabular}} &
  \textbf{\begin{tabular}[c]{@{}c@{}}81.22\\ 0.7359\end{tabular}} \\
\Xhline{5\arrayrulewidth}
\end{tabular}%
}
\caption{Experimental results of PiFi on a decoder-based model. For this experiment, we used SmolLM2-135B model with Llama-3.1-8B. We report accuracy and F1- score in the upper row and lower row of each cell.}
\label{tab:tab16-decoder model}
\end{table}\textbf{}

\subsection{Comparison between Text Representation Methods in Encoder-based LMs}
\label{app:ablation-representation}

In this section, we explore different methods for representing text in encoder-based LMs. In our original PiFi framework, we used the $\textit{CLS}$ token representation as $h_{\textit{enc}}$. To evaluate alternative approaches, we conducted experiments with two additional methods: (1) representing $h_{\textit{enc}}$ as the average of all token representations in the input sequence, including padding tokens, and (2) representing $h_{\textit{enc}}$ as the average of all token representations excluding padding tokens.

Table~\ref{tab:tab9-pooling} presents the experimental results. The results reveal distinct differences between these configurations, indicating that using the $\textit{CLS}$ token representation as $h_{\textit{enc}}$ is the most effective choice for ensuring optimal model performance.

% \usepackage{graphicx}
\begin{table}[t]
\centering
\resizebox{\columnwidth}{!}{%
\begin{tabular}{c|ccc}
\Xhline{3\arrayrulewidth}

\multicolumn{1}{l|}{\begin{tabular}[c]{@{}l@{}}BERT\textsubscript{\textit{base}}\\ \textbf{+ PiFi (Llama-3.1-8B)}\end{tabular}} &
  \textbf{SST-2} &
  \textbf{IMDB} &
  \begin{tabular}[c]{@{}c@{}}\textbf{Tweet}\\ \textbf{(Sentiment)}\end{tabular} \\ \hline\hline
CLS &
  \textbf{\begin{tabular}[c]{@{}c@{}}91.50\\ 0.9125\end{tabular}} &
  \textbf{\begin{tabular}[c]{@{}c@{}}87.09\\ 0.4800\end{tabular}} &
  \textbf{\begin{tabular}[c]{@{}c@{}}92.95\\ 0.9224\end{tabular}} \\ \hline
\begin{tabular}[c]{@{}c@{}}Avearge Pooling\\ (w/o padding)\end{tabular} &
  \begin{tabular}[c]{@{}c@{}}91.06\\ 0.9079\end{tabular} &
  \begin{tabular}[c]{@{}c@{}}86.25\\ 0.4724\end{tabular} &
  {\color[HTML]{363A3D} \begin{tabular}[c]{@{}c@{}}92.70\\ 0.9197\end{tabular}} \\ \hline
\begin{tabular}[c]{@{}c@{}}Avearge Pooling\\ (w/ padding)\end{tabular} &
  \begin{tabular}[c]{@{}c@{}}90.40\\ 0.9008\end{tabular} &
  \begin{tabular}[c]{@{}c@{}}86.56\\ 0.4758\end{tabular} &
  \begin{tabular}[c]{@{}c@{}}92.73\\ 0.9191\end{tabular}\\
\Xhline{3\arrayrulewidth}
  
\end{tabular}%
}
\caption{Experimental results comparing the method for representing the input sequence for encoder-based LMs with PiFi. For this experiment, we used BERT\textsubscript{\textit{base}} model with Llama-3.1-8B. We report accuracy and F1-score in the upper row and lower row of each cell.}
\label{tab:tab9-pooling}
\end{table}
% \usepackage{graphicx}
\begin{table}[t]
\centering
\resizebox{\columnwidth}{!}{%
\begin{tabular}{l|ccc}
\Xhline{3\arrayrulewidth}

 &
  \textbf{SST-2} &
  \textbf{IMDB} &
  \begin{tabular}[c]{@{}c@{}}\textbf{Tweet}\\ \textbf{(Sentiment)}\end{tabular} \\ \hline\hline
BERT\textsubscript{\textit{base}} &
  \begin{tabular}[c]{@{}c@{}}89.41\\ 0.8907\end{tabular} &
  \begin{tabular}[c]{@{}c@{}}85.10\\ 0.4733\end{tabular} &
  \begin{tabular}[c]{@{}c@{}}86.90\\ 0.862\end{tabular} \\ \hline
\begin{tabular}[c]{@{}l@{}}\textbf{+ PiFi}\\ \textbf{(Llama-3.1-8B)}\end{tabular} &
  {\color[HTML]{363A3D} \textbf{\begin{tabular}[c]{@{}c@{}}91.50\\ 0.9125\end{tabular}}} &
  \textbf{\begin{tabular}[c]{@{}c@{}}87.09\\ 0.4800\end{tabular}} &
  {\color[HTML]{363A3D} \textbf{\begin{tabular}[c]{@{}c@{}}92.95\\ 0.9224\end{tabular}}} \\ \hline
\begin{tabular}[c]{@{}l@{}}\textbf{+ PiFi-full}\\ \textbf{(Llama-3.1-8B)}\end{tabular} &
  {\color[HTML]{363A3D} \begin{tabular}[c]{@{}c@{}}91.27\\ 0.9106\end{tabular}} &
  \begin{tabular}[c]{@{}c@{}}86.00\\ 0.4726\end{tabular} &
  \begin{tabular}[c]{@{}c@{}}90.20\\ 0.8951\end{tabular}\\
\Xhline{3\arrayrulewidth}
  
\end{tabular}%
}
\caption{Experimental results to validate the effectiveness of fully fine-tuning $L_{\text{LLM}}$ rather than keeping it frozen. We report accuracy and F1-score in the upper row and lower row of each cell.}
\label{tab:tab10-fullft}
\end{table}

\subsection{Full fine-tuning of $L_{\textit{LLM}}$ in PiFi}
\label{app:ablation-full-ft}

In PiFi framework, we fine-tune the model while keeping $L_{\textit{LLM}}$ frozen. To validate the impact of this approach, we conducted an experiment where the entire PiFi model, including $L_{\textit{LLM}}$, was fully fine-tuned.

Table~\ref{tab:tab10-fullft} shows the results of this experiment. Freezing $L_{\textit{LLM}}$ during fine-tuning leads to better performance compared to fully fine-tuning the model. We hypothesize that freezing $L_{\textit{LLM}}$ preserves the extensive knowledge of the LLM, while full fine-tuning may cause catastrophic forgetting of this knowledge. Thus, it is crucial to keep $L_{\textit{LLM}}$ frozen to retain the LLM’s vast knowledge, optimize the SLM for the desired task, and reduce training costs.

% \usepackage{graphicx}
\begin{table}[t]
\centering
\resizebox{\columnwidth}{!}{%
\begin{tabular}{l|ccccc}
\Xhline{3\arrayrulewidth}
 &
  \textbf{SST-2} &
  \textbf{IMDB} &
  \begin{tabular}[c]{@{}c@{}}\textbf{Tweet}\\ \textbf{(Sentiment)}\end{tabular} &
  \begin{tabular}[c]{@{}c@{}}\textbf{Tweet}\\ \textbf{(Offensive)}\end{tabular} &
  \textbf{CoLA} \\ \hline\hline
BERT\textsubscript{\textit{base}} &
  \begin{tabular}[c]{@{}c@{}}89.41\\ 0.8907\end{tabular} &
  \begin{tabular}[c]{@{}c@{}}85.10\\ 0.4733\end{tabular} &
  \begin{tabular}[c]{@{}c@{}}86.90\\ 0.8620\end{tabular} &
  \begin{tabular}[c]{@{}c@{}}83.15\\ 0.7727\end{tabular} &
  \begin{tabular}[c]{@{}c@{}}80.10\\ 0.7398\end{tabular} \\\hline
\begin{tabular}[c]{@{}l@{}}\textbf{+ PiFi}\\ \textbf{(Llama-3.1-8B)}\end{tabular} &
  \begin{tabular}[c]{@{}c@{}}91.50\\ 0.9125\end{tabular} &
  \textbf{\begin{tabular}[c]{@{}c@{}}87.09\\ 0.4800\end{tabular}} &
  \textbf{\begin{tabular}[c]{@{}c@{}}92.95\\ 0.9224\end{tabular}} &
  \textbf{\begin{tabular}[c]{@{}c@{}}86.03\\ 0.8026\end{tabular}} &
  \begin{tabular}[c]{@{}c@{}}82.07\\ 0.7523\end{tabular} \\\hline
\begin{tabular}[c]{@{}l@{}}\textbf{+ PiFi}\\ \textbf{(Llama-3.1-8B-Instruct)}\end{tabular} &
  \textbf{\begin{tabular}[c]{@{}c@{}}91.98\\ 0.9174\end{tabular}} &
  \begin{tabular}[c]{@{}c@{}}86.72\\ 0.4882\end{tabular} &
  \begin{tabular}[c]{@{}c@{}}92.81\\ 0.9206\end{tabular} &
  \begin{tabular}[c]{@{}c@{}}85.57\\ 0.7985\end{tabular} &
  \textbf{\begin{tabular}[c]{@{}c@{}}82.45\\ 0.7450\end{tabular}} \\ \hline\hline
p-value &
  0.7352 &
  0.3237 &
  0.4422 &
  0.2129 &
  0.5308\\
\Xhline{3\arrayrulewidth}
\end{tabular}%
}
\caption{Experimental results comparing the usage of base LLM and instruction-tuned LLM for PiFi. For this experiment, we used Llama-3.1-8B and Llama-3.1-8B-Instruct. The p-value denotes the statistical significance between the distribution of accuracy of PiFi with Llama-3.1-8B and Llama-3.1-8B-Instruct. We report accuracy and F1-score in the upper row and lower row of each cell.}
\label{tab:tab7-instruction}
\end{table}

\subsection{Effectiveness of Instruction-tuned LLM for PiFi}
\label{app:ablation-instruction}

Recent innovations in LLMs involve additional training to enable these models to better follow instructions provided in prompts, leading to improved performance on various downstream tasks \cite{ouyang2022training}. In this section, we explore the impact of using instruction-tuned LLMs for extracting $L_{\textit{LLM}}$ within the PiFi framework. Specifically, we compare the performance of PiFi when using Llama-3.1-8B, our default model, against Llama-3.1-8B-Instruct, an instruction-tuned version of the Llama-3.1-8B.

The results are presented in Table~\ref{tab:tab7-instruction}. Our findings indicate that using instruction-tuned LLMs with PiFi can enhance the performance of SLMs, but the performance gains are marginal. In fact, there is no consistent trend showing that instruction-tuned models outperform their base counterparts across all datasets. To better understand this observation, we conducted a statistical significance test by comparing the results obtained from five different random seeds. The p-values for accuracy distributions between the instruction-tuned and base LLMs were greater than 0.05, indicating that these performance differences are not statistically significant.

This suggests that incorporating instruction-tuned LLMs does not necessarily lead to significant improvements for PiFi. Instead, it highlights that the intrinsic knowledge encoded within the LLMs is more crucial for PiFi's success than their capacity to follow human instructions.

\begin{table}[t]
\centering
\resizebox{\columnwidth}{!}{%
\begin{tabular}{l|ccccc}
\Xhline{5\arrayrulewidth}

                                                                & SST-2                                                           & IMDB                                                            & \begin{tabular}[c]{@{}c@{}}Tweet\\ (Sentiment)\end{tabular}     & \begin{tabular}[c]{@{}c@{}}Tweet\\ (Offensive)\end{tabular}     & CoLA                                                            \\ \hline\hline
BERT-base                                                       & \begin{tabular}[c]{@{}c@{}}89.41\\ 0.8907\end{tabular}          & \begin{tabular}[c]{@{}c@{}}85.10\\ 0.4733\end{tabular}          & \begin{tabular}[c]{@{}c@{}}86.90\\ 0.8620\end{tabular}          & \begin{tabular}[c]{@{}c@{}}83.15\\ 0.7727\end{tabular}          & \begin{tabular}[c]{@{}c@{}}80.77\\ 0.7296\end{tabular}          \\ \hline
\begin{tabular}[c]{@{}l@{}}+PiFi\\ (Llama-3.1-8B)\end{tabular}  & \textbf{\begin{tabular}[c]{@{}c@{}}91.50\\ 0.9125\end{tabular}} & \begin{tabular}[c]{@{}c@{}}87.09\\ 0.4800\end{tabular}          & \textbf{\begin{tabular}[c]{@{}c@{}}92.95\\ 0.9224\end{tabular}} & \textbf{\begin{tabular}[c]{@{}c@{}}86.03\\ 0.8026\end{tabular}} & \textbf{\begin{tabular}[c]{@{}c@{}}82.07\\ 0.7523\end{tabular}} \\ \hline
\begin{tabular}[c]{@{}l@{}}+PiFi\\ (Llama-3.1-70B)\end{tabular} & \begin{tabular}[c]{@{}c@{}}91.49\\ 0.9120\end{tabular}          & \textbf{\begin{tabular}[c]{@{}c@{}}87.11\\ 0.4875\end{tabular}} & \begin{tabular}[c]{@{}c@{}}92.73\\ 0.9198\end{tabular}          & \begin{tabular}[c]{@{}c@{}}85.00\\ 0.7784\end{tabular}          & \begin{tabular}[c]{@{}c@{}}82.00\\ 0.761\end{tabular}           \\ \hline
\begin{tabular}[c]{@{}l@{}}+PiFi\\ (Qwen2.5-7B)\end{tabular}    & \begin{tabular}[c]{@{}c@{}}91.05\\ 0.9080\end{tabular}          & \begin{tabular}[c]{@{}c@{}}86.56\\ 0.4765\end{tabular}          & \begin{tabular}[c]{@{}c@{}}91.90\\ 0.9117\end{tabular}          & \begin{tabular}[c]{@{}c@{}}85.14\\ 0.8007\end{tabular}          & \begin{tabular}[c]{@{}c@{}}81.65\\ 0.7562\end{tabular}          \\ \hline
\begin{tabular}[c]{@{}l@{}}+PiFi\\ (Qwen2.5-32B)\end{tabular}   & \begin{tabular}[c]{@{}c@{}}91.12\\ 0.9086\end{tabular}          & \begin{tabular}[c]{@{}c@{}}86.27\\ 0.4718\end{tabular}          & \begin{tabular}[c]{@{}c@{}}91.69\\ 0.9096\end{tabular}          & \begin{tabular}[c]{@{}c@{}}83.75\\ 0.7569\end{tabular}          & \begin{tabular}[c]{@{}c@{}}80.74\\ 0.7463\end{tabular}          \\ \hline
\begin{tabular}[c]{@{}l@{}}+PiFi\\ (Qwen2.5-72B)\end{tabular}   & \begin{tabular}[c]{@{}c@{}}91.38\\ 0.9115\end{tabular}          & \begin{tabular}[c]{@{}c@{}}86.75\\ 0.4788\end{tabular}          & \begin{tabular}[c]{@{}c@{}}92.21\\ 0.9150\end{tabular}          & \begin{tabular}[c]{@{}c@{}}84.51\\ 0.7796\end{tabular}          & \begin{tabular}[c]{@{}c@{}}82.45\\ 0.7450\end{tabular}         
\\ \Xhline{5\arrayrulewidth}
\end{tabular}
}
\caption{Experimental results comparing PiFi performance across various large language models. We evaluated PiFi with Llama-3.1 models (8B, 70B) and Qwen2.5 models (7B, 32B, 72B) across five downstream tasks. We report accuracy and F1-score in the upper row and lower row of each cell.}
\label{tab:tab21-large_llm}
\end{table}
% \usepackage{graphicx}
% \usepackage{graphicx}
\begin{table}[t]
\centering
\resizebox{\columnwidth}{!}{%
\begin{tabular}{l|ccccc}
\Xhline{5\arrayrulewidth}
 &
  SST-2 &
  IMDB &
  \begin{tabular}[c]{@{}c@{}}Tweet\\ (Sentiment)\end{tabular} &
  \begin{tabular}[c]{@{}c@{}}Tweet\\ (Offensive)\end{tabular} &
  CoLA \\ \hline\hline
BERT\textsubscript{\textit{base}} &
  \begin{tabular}[c]{@{}c@{}}89.41\\ 0.8907\end{tabular} &
  \begin{tabular}[c]{@{}c@{}}85.10\\ 0.4733\end{tabular} &
  \begin{tabular}[c]{@{}c@{}}86.90\\ 0.8620\end{tabular} &
  \begin{tabular}[c]{@{}c@{}}83.15\\ 0.7727\end{tabular} &
  \begin{tabular}[c]{@{}c@{}}80.77\\ 0.7296\end{tabular} \\ \hline
\begin{tabular}[c]{@{}l@{}}+PiFi\\ (Llama-3.1-8B)\end{tabular} &
  \begin{tabular}[c]{@{}c@{}}91.50\\ 0.9125\end{tabular} &
  \begin{tabular}[c]{@{}c@{}}87.09\\ 0.4800\end{tabular} &
  \textbf{\begin{tabular}[c]{@{}c@{}}92.95\\ 0.9224\end{tabular}} &
  \textbf{\begin{tabular}[c]{@{}c@{}}86.03\\ 0.8026\end{tabular}} &
  \begin{tabular}[c]{@{}c@{}}82.07\\ 0.7523\end{tabular} \\ \hline
\begin{tabular}[c]{@{}l@{}}+PiFi\\ (Llama-3.1-70B)\end{tabular} &
  \begin{tabular}[c]{@{}c@{}}91.49\\ 0.9120\end{tabular} &
  \begin{tabular}[c]{@{}c@{}}87.11\\ 0.4875\end{tabular} &
  \begin{tabular}[c]{@{}c@{}}92.73\\ 0.9198\end{tabular} &
  \begin{tabular}[c]{@{}c@{}}85.00\\ 0.7784\end{tabular} &
  \begin{tabular}[c]{@{}c@{}}82.00\\ 0.761\end{tabular} \\ \hline
\begin{tabular}[c]{@{}l@{}}+PiFi-2 layers\\ (Llama-3.1-70B)\end{tabular} &
  \textbf{\begin{tabular}[c]{@{}c@{}}91.76\\ 0.9154\end{tabular}} &
  \textbf{\begin{tabular}[c]{@{}c@{}}87.20\\ 0.4901\end{tabular}} &
  \begin{tabular}[c]{@{}c@{}}92.95\\ 0.9217\end{tabular} &
  \begin{tabular}[c]{@{}c@{}}85.66\\ 0.8119\end{tabular} &
  \textbf{\begin{tabular}[c]{@{}c@{}}82.22\\ 0.7488\end{tabular}}\\
\Xhline{5\arrayrulewidth}
  
\end{tabular}%
}
\caption{Experimental results of PiFi with multiple layers. This experiment compares the performance of using a single layer versus two layers from Llama-3.1-70B. We report accuracy in the upper row and F1-score in the lower row of each cell.}
\label{tab:tab11-70}
\end{table}

\subsection{Impact of Larger-Scale LLM on PiFi Performance}
\label{app:ablation-larger-scaleLLM}

Previously, as analyzed in Section 4.8, we examined the performance of PiFi using the Qwen model across different scales (0.5B, 1.5B, and 7B parameters). However, it is necessary to further examine how performance changes when applying larger-scale LLM layers. To this end, we conducted experiments applying PiFi to various large-scale models, including the 8B and 70B versions of Llama-3.1 and the 7B, 32B, and 72B versions of Qwen2.5. We selected Qwen2.5 as it was newly released at the time of writing this paper, allowing us to evaluate PiFi's scalability using state-of-the-art models with cutting-edge performance.

As shown in Table~\ref{tab:tab21-large_llm}, when comparing various large language models, performance improvements did not consistently correlate with model size. In particular, the use of the Llama-3.1-70B model layer in PiFi resulted in comparable or slightly lower performance than when using the 8B model layer. Similarly, the performance differences between Qwen2.5-7B, Qwen2.5-32B, and Qwen2.5-72B did not show a clear linear relationship with model size. We interpret this outcome as potentially influenced by a factor we refer to as layer knowledge density. Specifically, the 8B model comprises 32 layers, while the 70B model consists of 80 layers \cite{dubey2024llama}. This suggests that a single layer in the 70B model may encapsulate relatively less dense knowledge compared to a single layer in the 8B model.

To mitigate the limitations imposed by this lower density, we conducted additional experiments incorporating the last two layers of the 70B model, with results presented in Table~\ref{tab:tab11-70}. This experiment directly compares the performance when using a single layer versus two layers from the 70B model. However, the observed performance improvement was marginal. This suggests that the selected layer combination may not be optimal, and alternative configurations could lead to greater performance gains. Nevertheless, the large number of possible combinations poses a challenge in identifying the most effective configuration. Future research will explore more efficient methods for identifying effective layer combinations, aiming to enhance PiFi's scalability while optimizing performance.

% \usepackage{graphicx}
\begin{table}[t]
\centering
\resizebox{\columnwidth}{!}{%
\begin{tabular}{l|ccccc}
\Xhline{5\arrayrulewidth}

 &
  SST-2 &
  IMDB &
  \begin{tabular}[c]{@{}c@{}}Tweet\\ (Sentiment)\end{tabular} &
  \begin{tabular}[c]{@{}c@{}}Tweet\\ (Offensive)\end{tabular} &
  CoLA \\ \hline\hline
BERT\textsubscript{\textit{base}} &
  \begin{tabular}[c]{@{}c@{}}89.41\\ 0.8907\end{tabular} &
  \begin{tabular}[c]{@{}c@{}}85.10\\ 0.4733\end{tabular} &
  \begin{tabular}[c]{@{}c@{}}86.90\\ 0.8620\end{tabular} &
  \begin{tabular}[c]{@{}c@{}}83.15\\ 0.7727\end{tabular} &
  \begin{tabular}[c]{@{}c@{}}80.10\\ 0.7398\end{tabular} \\ \hline
\textbf{\begin{tabular}[c]{@{}l@{}}+PiFi\\ (Llama-3.1-8B)\end{tabular}} &
  \textbf{\begin{tabular}[c]{@{}c@{}}91.50\\ 0.9125\end{tabular}} &
  \textbf{\begin{tabular}[c]{@{}c@{}}87.09\\ 0.4800\end{tabular}} &
  \textbf{\begin{tabular}[c]{@{}c@{}}92.95\\ 0.9224\end{tabular}} &
  \textbf{\begin{tabular}[c]{@{}c@{}}86.03\\ 0.8026\end{tabular}} &
  \textbf{\begin{tabular}[c]{@{}c@{}}82.07\\ 0.7523\end{tabular}} \\ \hline
\begin{tabular}[c]{@{}l@{}}Single layer: First\\ (Llama-3.1-8B)\end{tabular} &
  \begin{tabular}[c]{@{}c@{}}80.09\\ 0.7970\end{tabular} &
  \begin{tabular}[c]{@{}c@{}}77.63\\ 0.4383\end{tabular} &
  \begin{tabular}[c]{@{}c@{}}80.14\\ 0.7913\end{tabular} &
  \begin{tabular}[c]{@{}c@{}}78.59\\ 0.6735\end{tabular} &
  \begin{tabular}[c]{@{}c@{}}68.12\\ 0.3812\end{tabular} \\ \hline
\begin{tabular}[c]{@{}l@{}}Single layer: Last\\ (Llama-3.1-8B)\end{tabular} &
  \begin{tabular}[c]{@{}c@{}}80.67\\ 0.8010\end{tabular} &
  \begin{tabular}[c]{@{}c@{}}78.31\\ 0.4392\end{tabular} &
  \begin{tabular}[c]{@{}c@{}}80.28\\ 0.7939\end{tabular} &
  \begin{tabular}[c]{@{}c@{}}80.47\\ 0.6983\end{tabular} &
  \begin{tabular}[c]{@{}c@{}}69.31\\ 0.4076\end{tabular} \\
\Xhline{5\arrayrulewidth}
  
\end{tabular}%
}
\caption{Experimental results on the usage of a single frozen $L_{\text{LLM}}$ without SLM. For this experiment, we extracted the first and last layers of Llama-3.1-8B as $L_{\text{LLM}}$. We report accuracy and F1-score in the upper row and lower row of each cell.}
\label{tab:tab12-single_layer}
\end{table}

\subsection{Evaluating the Usage of a Single LLM Layer}
\label{app:ablation-sigle_layer}
To evaluate the performance of a single frozen LLM Layer without SLM, we conducted experiments by attaching classification heads to the first and last layers of Llama-3.1-8B. 

As shown in the Table~\ref{tab:tab12-single_layer}, there is a clear performance gap between the PiFi framework and the approach using a single LLM Layer. This demonstrates that PiFi effectively enhances performance through integration with an SLM (e.g., BERT\textsubscript{\textit{base}}).

Notably, the single LLM Layer shows limitations in achieving high performance independently, highlighting the necessity of combining it with an SLM for effective knowledge transfer. Furthermore, despite having fewer parameters than PiFi, the single frozen LLM Layer exhibited significantly degraded performance across all tasks. These findings reaffirm that PiFi can enhance the performance of the SLM by effectively integrating knowledge from a single LLM Layer while preserving the efficiency of the SLM.

% \usepackage{graphicx}
\begin{table}[t]
\centering
\resizebox{\columnwidth}{!}{%
\begin{tabular}{l|c|c}
\Xhline{5\arrayrulewidth}
                                                                & FLOPs (GFLOPs) & GPU Memory (GB) \\ \hline\hline
\makebox[0pt][l]{\rule[-0.6em]{0pt}{2em}BERT\textsubscript{\textit{base}}}            & 272.12         & 1.33            \\ \hline
\textbf{\begin{tabular}[c]{@{}l@{}}+ PiFi\\ (Llama-3.1-8B)\end{tabular}} & 279.3          & 2.27            \\
\Xhline{5\arrayrulewidth}
\end{tabular}%
}
\caption{Efficiency-performance trade-off analysis results. This analysis was conducted by measuring FLOPs and GPU memory consumption.}
\label{tab:tab13-flops}
\end{table}

\subsection{Analyzing Efficiency-Performance Trade-offs in PiFi}
\label{app:ablation-flops}
As the number of parameters increases, it is necessary to analyze the trade-offs between efficiency and performance. To validate this, we measured FLOPs during the inference process for each model using the fvcore library \cite{Meta2019fvcore} and GPU memory consumption using the PyTorch Profiler \cite{paszke2019pytorchimperativestylehighperformance}. The results are presented in Table~\ref{tab:tab13-flops}.

The measurements revealed that PiFi adds approximately 2.6\% to the baseline inference cost in terms of FLOPs. By using only the CLS token as input to the $L_{\textit{LLM}}$, PiFi significantly reduces the sequence length processed in the $L_{\textit{LLM}}$s to a single token, thereby greatly reducing computational overhead. Although GPU memory consumption showed a relatively larger increase, this trade-off is deemed reasonable for tasks that require performance improvement in environments where additional memory resources are available. Moreover, PiFi’s modular design allows for flexible scalability according to system constraints, making it adaptable even in resource-constrained environments.

% \usepackage{graphicx}
\begin{table}[t]
\centering
\resizebox{\columnwidth}{!}{%
\begin{tabular}{l|ccccc}
\Xhline{5\arrayrulewidth}
 &
  SST-2 &
  IMDB &
  \begin{tabular}[c]{@{}c@{}}Tweet\\ (Sentiment)\end{tabular} &
  \begin{tabular}[c]{@{}c@{}}Tweet\\ (Offensive)\end{tabular} &
  CoLA \\ \hline\hline
BERT\textsubscript{\textit{base}} &
  \begin{tabular}[c]{@{}c@{}}89.41\\ 0.8907\end{tabular} &
  \begin{tabular}[c]{@{}c@{}}85.10\\ 0.4733\end{tabular} &
  \begin{tabular}[c]{@{}c@{}}86.90\\ 0.8620\end{tabular} &
  \begin{tabular}[c]{@{}c@{}}83.15\\ 0.7727\end{tabular} &
  \begin{tabular}[c]{@{}c@{}}80.10\\ 0.7398\end{tabular} \\ \hline
\textbf{\begin{tabular}[c]{@{}l@{}}+PiFi\\ (Llama-3.1-8B)\end{tabular}} &
  \textbf{\begin{tabular}[c]{@{}c@{}}91.50\\ 0.9125\end{tabular}} &
  \begin{tabular}[c]{@{}c@{}}87.09\\ 0.4800\end{tabular} &
  \textbf{\begin{tabular}[c]{@{}c@{}}92.95\\ 0.9224\end{tabular}} &
  \textbf{\begin{tabular}[c]{@{}c@{}}86.03\\ 0.8026\end{tabular}} &
  \textbf{\begin{tabular}[c]{@{}c@{}}82.07\\ 0.7523\end{tabular}} \\ \hline
\begin{tabular}[c]{@{}l@{}}BERT-base\\ + ZEROGEN (GPT2-XL)\end{tabular} &
  \begin{tabular}[c]{@{}c@{}}79.48\\ 0.7885\end{tabular} &
  \textbf{\begin{tabular}[c]{@{}c@{}}87.11\\ 0.4875\end{tabular}} &
  \begin{tabular}[c]{@{}c@{}}81.05\\ 0.7748\end{tabular} &
  \begin{tabular}[c]{@{}c@{}}27.89\\ 0.2146\end{tabular} &
  \begin{tabular}[c]{@{}c@{}}69.31\\ 0.4076\end{tabular} \\ \hline
\begin{tabular}[c]{@{}l@{}}BERT-base\\ + ZEROGEN (Llama-3.1-8B)\end{tabular} &
  \begin{tabular}[c]{@{}c@{}}81.56\\ 0.8106\end{tabular} &
  \begin{tabular}[c]{@{}c@{}}68.26\\ 0.4343\end{tabular} &
  \begin{tabular}[c]{@{}c@{}}87.99\\ 0.8637\end{tabular} &
  \begin{tabular}[c]{@{}c@{}}25.29\\ 0.2061\end{tabular} &
  \begin{tabular}[c]{@{}c@{}}30.69\\ 0.2306\end{tabular} \\
  \Xhline{5\arrayrulewidth}
\end{tabular}%
}
\caption{Experimental results comparing PiFi and ZEROGEN for knowledge transfer. For ZEROGEN, 200,000 synthetic examples were generated per domain using GPT-2-XL and Llama-3.1-8B as data generators. We report accuracy and F1-score in the upper row and lower row of each cell.}
\label{tab:tab14-zerogen}
\end{table}
\begin{table}[t]
\centering
\resizebox{\columnwidth}{!}{%
\begin{tabular}{l|ccccc}
\Xhline{5\arrayrulewidth}
                                                               & SST-2                                                           & IMDB                                                            & \begin{tabular}[c]{@{}c@{}}Tweet\\ (Sentiment)\end{tabular}     & \begin{tabular}[c]{@{}c@{}}Tweet\\ (Offensive)\end{tabular}     & CoLA                                                   \\ \hline \hline
BERT\textsubscript{\textit{base}}                                                      & \begin{tabular}[c]{@{}c@{}}89.41\\ 0.8907\end{tabular}          & \begin{tabular}[c]{@{}c@{}}85.10\\ 0.4733\end{tabular}          & \begin{tabular}[c]{@{}c@{}}86.90\\ 0.8620\end{tabular}          & \begin{tabular}[c]{@{}c@{}}83.15\\ 0.7727\end{tabular}          & \begin{tabular}[c]{@{}c@{}}80.77\\ 0.7296\end{tabular} \\ \hline
\begin{tabular}[c]{@{}l@{}}\textbf{+PiFi}\\ \textbf{(Llama-3.1-8B)}\end{tabular} & \textbf{\begin{tabular}[c]{@{}c@{}}91.50\\ 0.9125\end{tabular}} & \textbf{\begin{tabular}[c]{@{}c@{}}87.09\\ 0.4800\end{tabular}} & \textbf{\begin{tabular}[c]{@{}c@{}}92.95\\ 0.9224\end{tabular}} & \textbf{\begin{tabular}[c]{@{}c@{}}86.03\\ 0.8026\end{tabular}} & 
\textbf{\begin{tabular}[c]{@{}c@{}}82.07\\ 0.7523\end{tabular}} \\ \hline
BERT\textsubscript{\textit{base}}+LoRA                                                & \begin{tabular}[c]{@{}c@{}}82.93\\ 0.8207\end{tabular}          & \begin{tabular}[c]{@{}c@{}}75.22\\ 0.4287\end{tabular}          & \begin{tabular}[c]{@{}c@{}}81.37\\ 0.7677\end{tabular}          & \begin{tabular}[c]{@{}c@{}}73.07\\ 0.4490\end{tabular}          & \begin{tabular}[c]{@{}c@{}}69.98\\ 0.4252\end{tabular} \\ \Xhline{5\arrayrulewidth}
\end{tabular}
}
\caption{Experimental results comparing PiFi and LoRA for parameter-efficient fine-tuning. For LoRA, we used rank=8, scaling factor $\alpha$=16, and dropout=0.1 with the same BERT\textsubscript{\textit{base}} backbone. We report accuracy and F1-score in the upper row and lower row of each cell.}
\label{tab:tab20-lora}
\end{table}

\subsection{Comparison with Knowledge Distillation and Parameter-Efficient Fine-tuning Methods}
\label{app:ablation-zerogen}
We also conducted a comparison with ZEROGEN \cite{ye2022zerogen}, which leverages synthetic data generated by an LLM to enhance the SLM as an alternative method for distilling the inherent knowledge of an LLM. Following the settings of prior studies, we generated 200,000 synthetic examples per domain using GPT-2-XL. Additionally, to account for the fact that ZEROGEN's performance depends on the quality of the data generator, we conducted more comprehensive experiments by generating additional synthetic examples using the latest model, Llama-3.1-8B.

As shown in the Table~\ref{tab:tab14-zerogen}, ZEROGEN proved less effective than PiFi in most tasks, which we attribute to noise introduced during the synthetic data generation process. These results suggest that directly and efficiently utilizing the inherent knowledge of an LLM, rather than indirectly leveraging it through synthetic data generation, offers a more advantageous approach.

In addition to knowledge distillation methods, we compared PiFi with parameter-efficient fine-tuning using LoRA \cite{hu2022lora}, which has gained popularity for adapting language models with minimal computational overhead. For our experiments, we applied LoRA to the BERT-base backbone using standard settings (rank=8, scaling factor $\alpha$=16, dropout=0.1) and evaluated its performance on the same set of downstream tasks.

As shown in Table~\ref{tab:tab20-lora}, PiFi consistently outperforms LoRA across all downstream tasks. This demonstrates that PiFi achieves superior performance improvements by effectively integrating LLM layers, enabling direct transfer of rich linguistic knowledge and generalization capabilities encoded in large-scale models to SLMs.

% \usepackage{graphicx}
\begin{table}[t]
\centering
\resizebox{\columnwidth}{!}{%
\begin{tabular}{l|ccccc}
\Xhline{5\arrayrulewidth}
 &
  SST-2 &
  IMDB &
  \begin{tabular}[c]{@{}c@{}}Tweet\\ (Sentiment)\end{tabular} &
  \begin{tabular}[c]{@{}c@{}}Tweet\\ (Offensive)\end{tabular} &
  CoLA \\ \hline\hline
BERT\textsubscript{\textit{base}} &
  \begin{tabular}[c]{@{}c@{}}89.41\\ 0.8907\end{tabular} &
  \begin{tabular}[c]{@{}c@{}}85.10\\ 0.4733\end{tabular} &
  \begin{tabular}[c]{@{}c@{}}86.90\\ 0.8620\end{tabular} &
  \begin{tabular}[c]{@{}c@{}}83.15\\ 0.7727\end{tabular} &
  \begin{tabular}[c]{@{}c@{}}80.77\\ 0.7296\end{tabular} \\ \hline
\textbf{\begin{tabular}[c]{@{}l@{}}+PiFi\\ (Llama-3.1-8B)\end{tabular}} &
  \textbf{\begin{tabular}[c]{@{}c@{}}91.50\\ 0.9125\end{tabular}} &
  \textbf{\begin{tabular}[c]{@{}c@{}}87.09\\ 0.4800\end{tabular}} &
  \textbf{\begin{tabular}[c]{@{}c@{}}92.95\\ 0.9224\end{tabular}} &
  \textbf{\begin{tabular}[c]{@{}c@{}}86.03\\ 0.8026\end{tabular}} &
  \textbf{\begin{tabular}[c]{@{}c@{}}82.07\\ 0.7523\end{tabular}} \\ \hline
 &
  +2.09\%p &
  +1.99\%p &
  +6.05\%p &
  +2.88\%p &
  +1.97\%p \\ \hline\hline
BERT\textsubscript{\textit{base}}-10\% &
  \begin{tabular}[c]{@{}c@{}}87.32\\ 0.8691\end{tabular} &
  \begin{tabular}[c]{@{}c@{}}82.32\\ 0.4541\end{tabular} &
  \begin{tabular}[c]{@{}c@{}}85.40\\ 0.8463\end{tabular} &
  \begin{tabular}[c]{@{}c@{}}81.07\\ 0.7420\end{tabular} &
  \begin{tabular}[c]{@{}c@{}}75.14\\ 0.6866\end{tabular} \\ \hline
\textbf{\begin{tabular}[c]{@{}l@{}}+PiFi-10\%\\ (Llama-3.1-8B)\end{tabular}} &
  \textbf{\begin{tabular}[c]{@{}c@{}}89.96\\ 0.8967\end{tabular}} &
  \textbf{\begin{tabular}[c]{@{}c@{}}84.12\\ 0.4688\end{tabular}} &
  \textbf{\begin{tabular}[c]{@{}c@{}}91.21\\ 0.9030\end{tabular}} &
  \textbf{\begin{tabular}[c]{@{}c@{}}84.31\\ 0.7807\end{tabular}} &
  \textbf{\begin{tabular}[c]{@{}c@{}}77.20\\ 0.6527\end{tabular}} \\ \hline
 &
  +2.64\%p &
  +1.80\%p &
  +5.81\%p &
  +3.24\%p &
  +2.06\%p \\
  \Xhline{5\arrayrulewidth}
\end{tabular}%

}
\caption{Experimental results of PiFi under limited training data conditions. We used only 10\% of the total training data for each dataset, and report accuracy in the upper row of each cell and F1-score in the lower row. }
\label{tab:tab17-low-resource}
\end{table}
\subsection{Performance Evaluation of PiFi under Limited Training Data}
\label{app:ablation-low-resource}
To assess whether the PiFi framework operates effectively in a limited training data environment, experiments were conducted using only 10\% of the total training data for each dataset. Specifically, the following sample counts were used: SST-2 (6,920 $\rightarrow$ 692), IMDB (20,000 $\rightarrow$ 2,000), Tweet for sentiment classification (45,615 $\rightarrow$ 4,562), Tweet for offensive classification (11,916 $\rightarrow$ 1,192), and  CoLA (5,536 $\rightarrow$ 554).

As shown in Table~\ref{tab:tab17-low-resource}, even when trained on restricted data, the model with PiFi consistently outperformed the BERT\textsubscript{\textit{base}}. In particular, the SST-2, Tweet for offensive classification, and CoLA datasets exhibited relatively greater performance improvements compared to when the full dataset was used, indicating that PiFi can deliver significant benefits even under constrained data conditions.
In contrast, the IMDB and Tweet for sentiment classification datasets have a considerably larger amount of total data (20,000 and 45,615 samples, respectively), so even with only 10\% of the data, a sufficient number of training samples was available. Consequently, the relative performance gains were smaller; however, this can be viewed as a positive example of PiFi’s flexible applicability even without large-scale data. These results underscore that PiFi offers stable and consistent performance advantages in limited training data environments, thereby enhancing its potential for real-world low-resource scenarios.

% \usepackage{multirow}
% \usepackage{graphicx}
\begin{table*}[t]
\centering
\resizebox{\textwidth}{!}{%
\begin{tabular}{c|c|c|c|c|c}
\Xhline{3\arrayrulewidth}

\multicolumn{1}{l|}{} &
  \textbf{SLM Params(M)} &
  \textbf{$L_{\textit{LLM}}$ Params(M)} &
  \textbf{$L_{\textit{in}}$ Params(M)} &
  \textbf{$L_{\textit{out}}$ Params(M)} &
  \textbf{$\textit{Head}$ Params(M)} \\ \hline\hline
\multicolumn{1}{l|}{\begin{tabular}[c]{@{}l@{}}BERT\textsubscript{\textit{base}}\\ 
\textbf{+ PiFi (Llama-3.1-8B)}\end{tabular}}     & 109.482 &         &       &       &       \\ \cline{1-2}
\multicolumn{1}{l|}{\begin{tabular}[c]{@{}l@{}}RoBERTa\textsubscript{\textit{base}}\\ 
\textbf{+ PiFi (Llama-3.1-8B)}\end{tabular}}       & 124.646 &         &       &       &       \\ \cline{1-2}
\multicolumn{1}{l|}{\begin{tabular}[c]{@{}l@{}}ELECTRA\textsubscript{\textit{base}}\\ 
\textbf{+ PiFi (Llama-3.1-8B)}\end{tabular}}        & 108.892 &         &       &       &       \\ \cline{1-2}
\multicolumn{1}{l|}{\begin{tabular}[c]{@{}l@{}}DeBERTa\textsubscript{\textit{base}}\\ 
\textbf{+ PiFi (Llama-3.1-8B)}\end{tabular}}       & 138.602 & 218.112 & 3.146 & 3.146 &       \\ \cline{1-2}
\multicolumn{1}{l|}{\begin{tabular}[c]{@{}l@{}}DeBERTa-V3\textsubscript{\textit{base}}\\ 
\textbf{+ PiFi (Llama-3.1-8B)}\end{tabular}}   & 183.832 &         &       &       &       \\ \cline{1-2}
\multicolumn{1}{l|}{\begin{tabular}[c]{@{}l@{}}BERT\textsubscript{\textit{base}}\\ 
\textbf{+ PiFi (Mistral-7B-v0.1)}\end{tabular}} & 109.482 &         &       &       &       \\ \cline{1-2}
\multicolumn{1}{l|}{\begin{tabular}[c]{@{}l@{}}BERT\textsubscript{\textit{base}}\\ 
\textbf{+ PiFi (Mistral-7B-v0.3)}\end{tabular}}  & 109.482 &         &       &       & 0.592 \\ \cline{1-5}
\multicolumn{1}{l|}{\begin{tabular}[c]{@{}l@{}}BERT\textsubscript{\textit{base}}\\ 
\textbf{+ PiFi (Qwen2-0.5B)}\end{tabular}} 
     &         & 14.912  & 0.688 & 0.688 &       \\ \cline{1-1} \cline{3-5}
\multicolumn{1}{l|}{\begin{tabular}[c]{@{}l@{}}BERT\textsubscript{\textit{base}}\\ 
\textbf{+ PiFi (Qwen2-1.5B)}\end{tabular}} 
     &         & 46.798  & 1.180 & 1.180 &       \\ \cline{1-1} \cline{3-5}
\multicolumn{1}{l|}{\begin{tabular}[c]{@{}l@{}}BERT\textsubscript{\textit{base}}\\ 
\textbf{+ PiFi (Qwen2-7B)}\end{tabular}}  & \multirow{3}{*}{109.482} & 233.058 & 3.146 & 3.146 &       \\ \cline{1-1} \cline{3-5}
\multicolumn{1}{l|}{\begin{tabular}[c]{@{}l@{}}BERT\textsubscript{\textit{base}}\\ 
\textbf{+ PiFi (Gemma-2-9B)}\end{tabular}} &         & 198.195 & 3.146 & 3.146 &       \\ \cline{1-1} \cline{3-5}
\multicolumn{1}{l|}{\begin{tabular}[c]{@{}l@{}}BERT\textsubscript{\textit{base}}\\ 
\textbf{+ PiFi (Falcon-7B)}\end{tabular}} &         & 207.070 & 3.146 & 3.146 &       \\ \cline{1-1} \cline{3-5}
\multicolumn{1}{l|}{\begin{tabular}[c]{@{}l@{}}BERT\textsubscript{\textit{base}}\\ 
\textbf{+ PiFi (Llama-3.1-70B)}\end{tabular}} &         & 855.654 & 6.291 & 6.291 &       \\ \cline{1-6}
\multicolumn{1}{l|}{\begin{tabular}[c]{@{}l@{}}T5\textsubscript{\textit{base}}\\ \textbf{+ PiFi (Llama-3.1-8B)}\end{tabular}}   &
  222.904 &
  \multirow{2}*[-2.5ex]{\centering 218.112} &
  \multirow{2}*[-2.5ex]{\centering 3.146} &
  \multirow{2}*[-2.5ex]{\centering 3.146} &
  \multirow{2}*[-2.5ex]{\centering -} \\ \cline{1-2}
\multicolumn{1}{l|}{\begin{tabular}[c]{@{}l@{}}BART\textsubscript{\textit{base}}\\ \textbf{+ PiFi (Llama-3.1-8B)}\end{tabular}}   & 139.420 &         &       &       &    \\
\Xhline{3\arrayrulewidth}
\end{tabular}%
}
\caption{The parameters of each module for PiFi, across various setup regarding SLMs and LLMs.}
\label{tab:tab8-parameter}
\end{table*}

\end{document}